\documentclass[letterpaper]{article} 
\usepackage[preprint]{aaai2027}  

\usepackage[hyphens]{url}  
\usepackage{graphicx} 
\usepackage{natbib}  
\usepackage{caption} 
\usepackage{algorithm}
\usepackage{algorithmic}

\usepackage{booktabs}
\usepackage{multirow}

\usepackage{amsmath}
\usepackage{amssymb}
\usepackage{mathtools}
\usepackage{amsthm}

\usepackage{enumitem}

\usepackage{subcaption}

\newcommand{\zH}{z_\mathrm{H}}
\newcommand{\zL}{z_\mathrm{L}}

\newcommand{\blfootnote}[1]{%
  \begingroup
  \renewcommand{\thefootnote}{}%
  \footnote{#1}%
  \addtocounter{footnote}{-1}%
  \endgroup
}

\title{Dissecting Hierarchical Reasoning Models: A Mechanistic Study}

\author{
    Leo Raphael Rodrigues,
    Jian Kang
}
\affiliations{
Mohamed bin Zayed University of Artificial Intelligence\\
}

\begin{document}

\maketitle

\blfootnote{%
  \textit{Correspondence to:}
  \{Leo.Rodrigues,Jian.Kang\}@mbzuai.ac.ae.
  \quad
  \textit{Github repository for code:}
  \url{https://github.com/LeoRodrigues05/HRM}
}

\begin{abstract}
Latent-space reasoning models can perform iterative computation without verbalizing each intermediate step, making it difficult to determine which internal states support prediction and whether decodable features correspond to mechanisms used during inference. We study Hierarchical Reasoning Model (HRM), a representative hierarchical Transformer-based reasoning model with many variants, on Sudoku, Maze, and ARC-AGI-2. We mechanistically understand how HRM reasons and what information it encodes. 
Our analyses compare HRM against Transformer baselines with and without recurrent modules, apply causal interventions on recurrent states, and utilize linear probes against random-direction ablations, as well as sparse autoencoders with feature ablations. Our results reveal several key findings: recurrent models outperform one-pass baselines, while single-state recurrent Transformers are comparable to HRM. State interventions further show that the causal contributions of the high- and low-level states vary across task-specific checkpoints and inference stages. Selected task variables are linearly decodable from the recurrent states in HRM, yet ablating probe directions produce effects comparable to random controls. SAE ablations yield larger behavioral changes than probe-direction ablations. However, top-ranked SAE features show no stable advantage over size-matched random subsets at larger ablation sizes or across tasks; the same pattern persists in a Sudoku control with within-step BPTT. Together, we characterize that HRM is essentially implementing constraint-aware iterative refinement on a puzzle-specific solution state, in which the functional contributions of components at different levels vary without relying on a compact, causally important feature set. These results highlight the necessity of studying the different working mechanisms and the importance of developing mechanistic interpretability techniques better suited for latent-space, recursive reasoning models.


\end{abstract}
\section{Introduction}~\label{sec:intro}
Reasoning can help solve problems whose solutions cannot be obtained through simple pattern matching~\citep{wei2023chainofthoughtpromptingelicitsreasoning, bubeck2023sparksagi} and is widely applied in mathematical proofs~\citep{trinh2024olympiad}, software development~\citep{jimenez2024swebench}, and scientific discovery~\citep{jumper2021alphafold}.
Different from reasoning in large language models that verbalizes an explicit chain of natural language intermediate steps~\citep{wei2023chainofthoughtpromptingelicitsreasoning}, latent-space reasoning offers a complementary route, which simply refines its continuous latent states directly~\citep{hao2025coconut, wang2025hrm, jolicoeur2025tiny, qasim2025acceleratingtrainingspeedtiny, baek2026generativerecursivereasoning}. 

These systems show that internal reasoning can be expanded without a verbal reasoning trace. This makes latent-space reasoning a practical design for constraint satisfaction, spatial planning, and other problems whose intermediate operations may be costly or unnatural to express as text \citep{yang2023learningsolveconstraintsatisfaction,tamar2017valueiterationnetworks,schwarzschild2021learnalgorithmgeneralizingeasy}. One way is to recycle latent states as intermediate inputs for latent chain-of-continuous-thoughts~\citep{hao2025coconut}. Recent latent-space reasoning models, on the other hand, repeatedly refine persistent states through shared computation in the Transformer block. The Hierarchical Reasoning Model (HRM)~\citep{wang2025hrm} is among the earliest and has inspired a growing family of latent reasoning recursive models, including the TRM~\citep{jolicoeur2025tiny}, CGAR~\citep{qasim2025acceleratingtrainingspeedtiny}, and Generative Recursive reasoning models~\citep{baek2026generativerecursivereasoning}. Analogous to fast and slow cognitive thinking~\citep{muarry2014hierarchy}, it works by coupling two recurrent modules at different timescales: a slow high-level state updated per cycle and a fast low-level state updated multiple times per cycle. Such a hierarchical nature makes HRM empirically strong on reasoning tasks like Sudoku, maze navigation, and ARC-AGI, using a small number of parameters (27M parameters only).

The property that makes HRM and latent-space reasoning appealing also removes its most accessible audit trail. End-task accuracy cannot reveal how the computation is divided between the high-level module and low-level module, which algorithmic strategy it uses to solve a reasoning task, and whether a readable feature is actually used. \citet{ren2026guessing} analyze whole-segment dynamics while abstracting away the inner high/low structure; \citet{miyanishi2026interaction} measures the spatial reach of high- and low-level perturbations in Sudoku, maze navigation, and ARC-AGI. We study complementary questions about state necessity across recurrent steps and the causal status of readable and sparse features.  Specifically, we ask:

\begin{itemize}[leftmargin=*]
    \item (RQ1) What functional roles do the high- and low-level states of HRM play during recurrent inference? Do the roles differ in different tasks?
    \item (RQ2) When task-relevant information is linearly decodable from the latent states of HRM, do the corresponding probe directions causally support inference?
    \item (RQ3) Is there a compact feature set from the features identified by probes and sparse autoencoders (SAEs) that impacts HRM performance the most?
\end{itemize}

To answer these questions, we explore Sudoku, Maze, and ARC-AGI using HRM. Specifically, we compare HRM against four baseline models based on Transformers with and without a recurrent module, apply causal interventions on latent representations across puzzles and steps, and conduct representational analysis using probes and SAEs~\citep{cunningham2024sparse} for feature discovery and causal validation of features. From the analyses, we have the following key findings:

\begin{itemize}[leftmargin=*]
    \item \textbf{HRM iteratively refines a puzzle-specific solution state with task-dependent, varying functional contributions of high- and low-level modules.} In Section~\ref{sec:finding1}, the solution state resides in the high-level activations on Sudoku, while it resides in the low level on Maze since the task largely saturates after one high-level update.

    \item \textbf{Linearly decodable probe readout $\neq$ causally relevant information.} In Section~\ref{sec:finding2}, constraints in Sudoku, Maze, and ARC-AGI tasks are linearly decodable from the high-level state at ${\sim}90\%$ accuracy. Yet, directed ablation of probe directions is not distinguishable from random controls. Readout, therefore, reveals the representational ability of features and not the mechanism in the model.

    \item \textbf{The probe and SAE features do not consistently isolate a compact set of causally dominant features.} In Section~\ref{sec:finding3}, our results show that ablation of SAE features trained on HRM activations produces significantly larger effects than linear probe directions, but the top 50 SAE features are no more impactful than 50 random features. This supports a distributed-computation view of HRM in which the causally necessary state of HRM is spread across the activation space, not localized by sparse feature bases.
\end{itemize}

\section{Related Work}~\label{sec:related}


\noindent \textbf{Reasoning in the latent space.}
Recent works in latent-space reasoning can be categorized into latent chain-of-thought (CoT) and recurrence in reasoning. For latent CoT, Coconut \citep{hao2025coconut} feeds the hidden state of a model back into itself as the next input embedding, mimicking the chain of thought in the discrete token space of large language models \citep{wei2023chainofthoughtpromptingelicitsreasoning}. Heima~\citep{shen2025efficientreasoninghiddenthinking} further compresses the hidden representations into shorter latent sequences, while \citet{zhou2025geometry} optimizes the geometry or length of latent thinking. These methods treat the latent state as a direct substitute for CoT tokens. Different from latent CoT, recurrent reasoning obtains computational depth in the latent space by repeatedly passing through a single neural network block over multiple iterations.
Adaptive Computation Time (ACT)~\citep{graves2016act} and Universal Transformers~\citep{dehghani2019universal} are early examples of recurrent reasoning. Latent-space recurrent reasoning is later scaled up by \citet{geiping2025recurrentdepth}. Hierarchical Reasoning Model (HRM)~\citep{wang2025hrm} reasons in the latent space by introducing a high-level module and a low-level module and letting these two modules interact. Related approaches further simplify recursive refinement with TRM~\citep{jolicoeur2025tiny, sghaier2026probabilistictiny}, introduce curriculum-guided recursive training~\citep{qasim2025acceleratingtrainingspeedtiny}, or model stochastic latent trajectories~\citep{baek2026generativerecursivereasoning}.
Recursive language models~\citep{zhang2025recursive} take recursion as a primary driver of computational depth for auto-regressive language modeling. These works aim to understand whether recursion helps improve reasoning capability and report their end-task accuracies. Our work primarily considers HRM and aims to understand how hierarchy and recurrence enable stronger reasoning performance with fewer parameters.

\noindent \textbf{Mechanistic interpretability.}
Mechanistic interpretability aims to understand the internal mechanisms of the learning model. One line of work studies how to decode interpretable features from the latent representations. Linear probes isolate what is linearly decodable from a model representation~\citep{belinkov2022probing}; while iterative null-space projection and amnesic probing extend probing with a causal experiment by ablating the probed direction and measuring downstream behavior~\citep{ravfogel2020null,elazar2021amnesic}. Sparse autoencoders (SAEs) decompose hidden states in an over-complete dictionary of sparsely activating features and have produced mono-semantic features in language models~\citep{bricken2023monosemanticity, cunningham2024sparse, templeton2024scaling}, with recent extensions to reasoning traces~\citep{chen2025sae_cot, theodorus2025sae}. Activation patching and trajectory analysis are alternatives to understanding how the model works internally. For example, 
\citet{zhang2025causal} performs causal analysis of continuous thought representations, \citet{vilas2025latent} characterizes latent temporal signals, \citet{zhou2025geometry} studies the geometry of reasoning trajectories, and \citet{bogdan2025thought} identifies thought anchors in the chain of thoughts. 
The works that are most relevant to ours are done by \citet{ren2026guessing} and \citet{miyanishi2026interaction}. Specifically, \citet{ren2026guessing} argues that HRM is a structured guesser through findings regarding fixed-point violations, latent dynamics across steps, and PCA-based reasoning modes, and proposes an ensemble-voting mechanism to improve Sudoku accuracy, but does not analyze what the hierarchical structure encodes nor test whether identified features are causally relevant during inference, which we address in this work. \citet{miyanishi2026interaction} introduces interaction locality and uses SAE feature ablation and finite-noise activation patching to study how local effects accumulate across HRM and TRM on Sudoku, Maze, and ARC-AGI. Complementary to this, we study the task dependent functional roles of HRMs high- and low-level activation states and test whether probe and SAE identified features localize their causal computation.

\section{Preliminaries and Setup}~\label{sec:background}
\noindent \textbf{Problem and settings.}
We consider supervised grid-to-grid reasoning problems. Representative problems include Sudoku, Maze and ARC-AGI. Specifically, 
\begin{itemize}[leftmargin=*]
    \item Sudoku is a $9 \times 9$ constraint-based puzzle, where blank cells must be filled with a digit from $1$ -- $9$ such that no digit is repeated in any row, column, or $3 \times 3$ sub-grid. We use the Sudoku-Extreme dataset that includes puzzles with difficulty ``extreme'' and the publicly released HRM checkpoint without retraining~\citep{wang2025hrm}. The input $\mathbf{x}$ is the flattened grid in row-major order with blanks encoded as a special token \texttt{<blank>}, and the flattened solution is represented as $\mathbf{y}$. A deterministic set of 500 puzzles are used for evaluation. We report cell accuracy, puzzle accuracy, Hamming distance to the ground truth, and the total number of row/column/sub-grid violations.
    
    \item Maze is a $30 \times 30$ grid path-finding task. The input grid encodes blocked, free, start, and goal cells; while the target grid marks a valid path from start to goal cells while preserving the initial layout. We use the Maze-Hard dataset and publicly available checkpoints without retraining~\citep{wang2025hrm}. We report path-cell accuracy, path precision/recall/F1/Jaccard, whether the predicted path connects the start and goal cells, avoids walls, has no branches, and if the path is valid and/or optimal. 

    \item ARC-AGI-2 consists of heterogeneous grid-transformation tasks, where a model observes input-output examples and predicts the transformed output grid on a held-out test input. Tasks vary in color, vocabulary, grid shape, output size, and transformation rule. We rebuild the ARC-2 augmented dataset~\citep{wang2025hrm} with random seed 42 and 1000 augmentations (e.g., rotate/reflect a grid, permutate colors).
    Starting from the publicly released checkpoint, we fine-tune a model using our rebuilt dataset with a smaller subset of puzzles and a smaller epoch budget. We report color-cell accuracy, exact solved, shape correctness, number-of-colors correctness, and color-Intersection over Union (color-IoU). Due to the heterogeneous nature of ARC-AGI tasks, it is hard to interpret activation patching and ablation. Thus, we only perform ARC-AGI for representation-level analysis using linear probes and SAEs.
\end{itemize}
For each problem, we provide dataset and evaluation details in Appendix~\ref{app:dataset}.

\noindent \textbf{Hierarchical Reasoning Model.} Hierarchical Reasoning Model (HRM) is a compact 27M-parameter recurrent encoder with two interacting Transformer modules: a low-level module $f_{\rm L}$ and a high-level module $f_{\rm H}$ that maintain carry states $\zL$ and $\zH$ respectively. Let $\mathbf{x}$ be the input. We first obtain an input embedding through an input head $\tilde{\mathbf{x}} = f_{\rm I}\left(\mathbf{x}\right)$ Within one cycle, the low-level state is updated $T$ times 
\begin{equation}
    \zL \leftarrow f_{\rm L}\left(\zL, \zH + \mathbf{x}\right)
\end{equation}
after which the high-level state is updated once
\begin{equation}
    \zH \leftarrow f_{\rm H}\left(\zH, \zL\right)
\end{equation}
A forward pass (i.e., segment) of HRM includes $N$ such cycles. Then $\zH$ will be projected to per-cell logits through an output head $f_{\rm O}$. After each forward pass, a learned adaptive-computation-time (ACT) head determines whether computation halts or continues~\citep{graves2016act}, subject to a maximum of $M_{\rm max} = 16$ segments. HRM is trained with deep supervision and a one-step gradient approximation (i.e., only the final $\zL$ and $\zH$ updates are differentiated). Unless stated otherwise, we set the hidden dimension $d = 512$ for both $\zL$ and $\zH$, and our analyses run all 16 segments to obtain a full $\zH^{\left(0\right)}, \ldots, \zH^{\left(15\right)}$ trajectory. Full update equations, training details, and the carry mechanism are in Appendix~\ref{app:hrm}.

\section{Finding 1: Functional Roles of HRM Components Vary across Tasks}
\label{sec:finding1}

We characterize the functional roles of high- and low-level states in HRM and test whether these roles are task independent or not. 
We primarily evaluate on Sudoku and Maze, since the heterogeneous nature of ARC-AGI-2 makes interpretation of activating patching and ablation challenging.

\begin{figure}[t]
\centering
\includegraphics[width=.8\columnwidth]{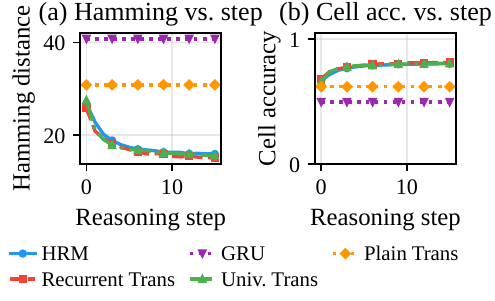}
\caption{Across all five architectures, recurrent models (HRM, Recurrent/Universal Transformer) \textbf{(a)} steadily reduce Hamming distance to the ground-truth solution and \textbf{(b)} improve cell accuracy over 16 ACT steps, while the non-recurrent baselines (Plain Transformer, GRU) remain flat at ${>}30$ mismatches and substantially lower accuracy.}
\label{fig:recurrence_curves}
\end{figure}

\paragraph{Recurrence is necessary.} To understand the contributions of \textit{recurrence} (i.e., reusing weights or hidden states across multiple refinement updates) and \textit{hierarchy} (i.e., high/low levels), we compare HRM against four baseline methods~\citep{knoop2025hrmdrivers}. The Recurrent Transformer updates an 8-layer post-norm Transformer for 16 steps, while the Universal Transformer~\citep{dehghani2019universal} is a weight-shared, single-state recurrent variant. The plain Transformer makes one forward pass, and GRU~\citep{cho2014gru} is recurrent over the input sequence but does not iteratively refine the same problem state. All models share the same input embedding and output head as HRM. The baselines use the same Sudoku data and broadly matched training settings, model components, and parameter counts~\citep{wang2025hrm}. We do not retrain HRM and evaluate all models on Sudoku-Extreme. Full configurations are provided in Appendix~\ref{app:baseline_recurrence} and ~\ref{app:baselines}.

Figure~\ref{fig:recurrence_curves} shows that HRM, Recurrent Transformer, and Universal Transformer progressively reduce Hamming distance and increase cell accuracy over 16 steps. At the final step, models with recurrence achieve higher cell accuracy and puzzle accuracy than the Transformer and GRU without recurrence. Nevertheless, Recurrent Transformer slightly outperforms HRM on puzzle accuracy, indicating that \textit{hierarchy does not itself clearly improve Sudoku at this scale.} HRM exhibits a different internal structure with a persistent high-level state $\zH$ amenable to targeted causal intervention.

\begin{figure}[t]
\centering
\includegraphics[width=\linewidth]{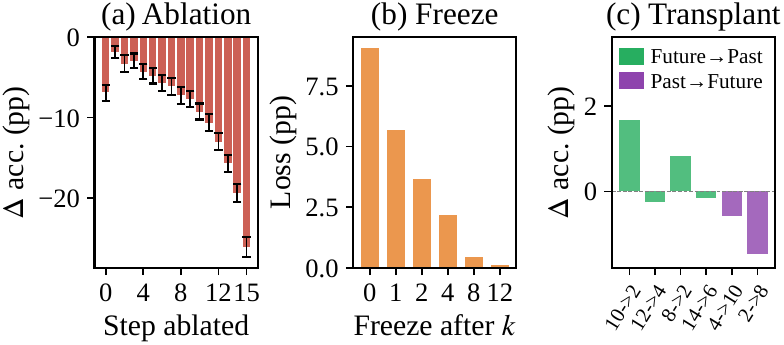}
\caption{Causal evidence for $\zH$ as a dynamic plan. \textbf{(a)}~Per-step ablation: zeroing $\zH$ is increasingly destructive at later steps ($-6.8$ at step~0 vs.\ $-26.0$ at step~15). \textbf{(b)}~Freeze curve: preventing $\zH$ updates after step $k$ shows progressive refinement, freezing at step~0 loses $-9.1$; by step~8, the plan is nearly complete. \textbf{(c)}~Time-shift: transplanting later $\zH$ states into earlier positions slightly \emph{improves} accuracy, confirming monotonic refinement.}
\label{fig:causal_evidence}
\end{figure}

\paragraph{Sudoku: $\zH$ is a progressively refined solution state.} We test whether $\zH$ carries the evolving solution by ablating $\zH$ or $\zL$ at selected inference steps over 1,000 puzzles. We find that ablating $\zH$ at all steps reduces cell accuracy significantly (-19.3\%), with per-step effect increasing (Figure~\ref{fig:causal_evidence}(a)). Although all-step $\zL$ ablation produces a similar aggregate effect, individual $\zL$ ablations are substantially weaker, indicating that the solution information is concentrated in $\zH$.

State freezing and within-instance temporal transplant provide further evidence. In Figure~\ref{fig:causal_evidence}(b), freezing $\zH$ from step 1 onward reduces accuracy by 9.1\%, with smaller effects at later freeze points. Freezing $\zL$ has negligible effects, consistent with a transient working-state role. In Figure~\ref{fig:causal_evidence}(c), transplanting a later $\zH$ to an earlier step slightly improves accuracy, while reducing accuracy vice versa. We also perform cross-puzzle patching to show that $\zH$ encodes puzzle-specific information. Replacing the target $\zH$ with one from a donor shifts the output toward the donor constraints, partially rewriting the given cells of the target. 

We provide full results in Appendix~\ref{app:interventions}. Together, these interventions identify $\zH$ as the solution state for Sudoku.

\paragraph{Maze: Patchable path state shifts to $\zL$.} Maze exhibits a different profile. Solutions largely stabilize after the first recurrent update. Though path-cell token accuracy changes modestly for all-step $\zH$ and $\zL$ ablations, path-based metrics reveal substantially larger causal effects. All-step $\zH$ ablations and per-step $\zH$ ablations both significantly decreases the proportion of valid paths, compared to ablating $\zL$, suggesting that $\zH$ remains important for producing the final path.

Cross-puzzle patching identifies a complementary role for $\zL$. Donor $\zL$ patching reduces valid-path rate by 43\%--56\% from step 4 onward, whereas patching $\zH$ has little effect until the final readout. Thus, $\zL$ carries intermediate path representation, while $\zH$ becomes most consequential at output.

These findings are consistent with the structure of Maze: the model receives a fully observed spatial graph and can form most of the route during the initial recurrent update.

\paragraph{Summary.} Together, these results answer RQ1: the hierarchy inside HRM provides two interpretable, intervenable, and separable recurrent modes. These hierarchical structure iteratively refines the solution across steps. However, the functional contributions vary across tasks, with $\zH$ carrying the solution state for Sudoku and $\zL$ for Maze. We also provide population-level analysis of Sudoku and Maze solution state trajectories in Appendix~\ref{app:traj_pca}.

\begin{figure}[t]
\centering
\includegraphics[width=\linewidth]{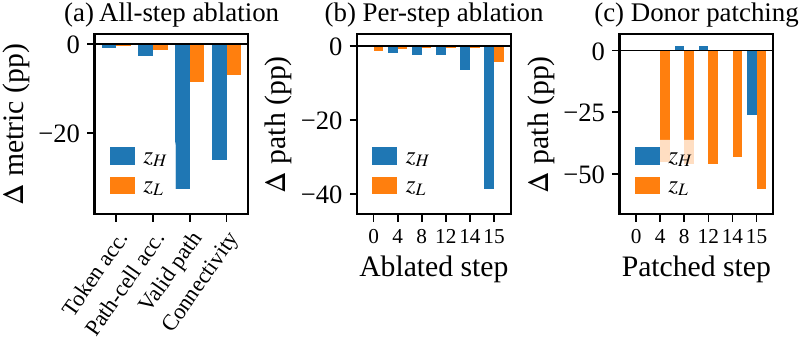}
\caption{Causal evidence for Maze's reorganized division of labor; values are percentage-point changes. \textbf{(a)}~All-step ablation: token accuracy moves little for either stream, but valid paths rate fall $-32.5\%$ under $\zH$ and $-8.5\%$ under $\zL$, so token accuracy is a weak diagnostic. \textbf{(b)}~Per-step ablation: $\zH$ matters the most at the final readout; no single-step $\zL$ ablation exceeds $-4.5$. \textbf{(c)}~Donor patching: donor-$\zL$ destroys $43$--$56\%$ of valid paths from step~4 onward, whereas donor-$\zH$ is inert until readout ($-26$ at step~15). $\zL$ is therefore the strongest patchable carrier of the intermediate path state.}
\label{fig}
\end{figure}

\section{Finding 2: Probe Readout $\neq$ Causally Relevant Information}~\label{sec:finding2}
Finding~1 establishes where the task specific solution state is causally relevant. RQ2 asks whether task variables can be decoded from the latent states and how causally important they are~\citep{hewitt2019designing, elazar2021amnesic, kumar2023probing}. Across Sudoku, Maze, and ARC-AGI-2, we find that many task variables are linearly decodable from HRM activations. But directed ablations of these identified features show weak causality or comparable effects like random controls to the end-task performance. Thus, probe accuracy only measures the accessibility of information but does not identify model's causal mechanism by itself.

\paragraph{Setup.} For each task, we record $\zH$ and $\zL$ at each recurrent step and train independent probes on held-out samples. Sudoku targets include cell identity, clue status, per-cell correctness, and row/column/sub-grid violations. Maze targets include local cell properties, such as walls, free cells, and path membership, as well as global properties including start-goal connectivity and path validity. ARC-AGI-2 targets describe grid geometry, color structure, object boundaries, and input-output transformations. We primarily use linear probes and report accuracy for each target. We provide complete target definitions in Appendix~\ref{app:probes}.

\begin{table*}[t]
\centering
\caption{Selected linear-probe validation accuracies across tasks. Sudoku and Maze results use final-step activations from both recurrent streams where available. ARC-AGI-2 is included as a representation-level replication using $\zH$ at step~15.
Expanded probe sweeps, MLP comparisons and complete target definitions are reported in Appendix~\ref{app:probes},~\ref{app:full_probes}.}
\label{tab:cross_task_probes}
\vskip 0.1in
\small
\begin{tabular}{@{}lllcc@{}}
\toprule
\textbf{Task} & \textbf{State / step} & \textbf{Target} &
\textbf{Probe acc.} & \textbf{Baseline / note} \\
\midrule
Sudoku & $\zH$, step~15 & Row / col / box violation &
88.3 / 87.8 / 87.5 & per-cell binary \\
Sudoku & $\zL$, step~15 & Row / col / box violation &
88.2 / 87.6 / 87.1 & per-cell binary \\
Sudoku & $\zH$, step~15 & Per-cell correctness &
82.5 & per-cell binary \\
Sudoku & $\zL$, step~15 & Per-cell correctness &
83.3 & per-cell binary \\
\midrule
Maze & $\zH$, step~15 & Exact solved / valid S--G path &
77.5 / 95.7 & global path features \\
Maze & $\zL$, step~15 & Exact solved / valid S--G path &
75.8 / 92.1 & global path features \\
Maze & $\zH$, step~15 & Connects start--goal / optimal path &
94.5 / 86.6 & global path features \\
Maze & $\zL$, step~15 & Connects start--goal / optimal path &
92.3 / 84.9 & global path features \\
\midrule
ARC-AGI-2 & $\zH$, step~15 & Per-cell correct / color changed &
91.6 / 97.2 & representation-level \\
ARC-AGI-2 & $\zH$, step~15 & Same as input / object boundary &
97.6 / 82.0 & representation-level \\
ARC-AGI-2 & $\zH$, step~15 & Input color / output color &
81.4 / 88.6 & 10-way multiclass \\
\bottomrule
\end{tabular}
\end{table*}

\paragraph{Linear probes decode constraint information.} Target variables are linearly predictable from both $\zH$ and $\zL$. \textit{Static, input-derived features} can be decoded from the earliest recurrent states. These features include Sudoku clue status and cell identity, Maze walls and free cells, and ARC-AGI-2 input colors and grid boundaries. \textit{Dynamic, solution-dependent features} become increasingly accessible over recurrent computation, with ${\sim}90\%$ accuracy at the final step as shown in Table~\ref{tab:cross_task_probes}. These include Sudoku constraint violations and per-cell correctness, Maze path validity and start-goal connectivity, and ARC-AGI-2 transformation-related features. Detailed per-step probe sweeps for both linear probes and MLP probes are provided in Appendix~\ref{app:full_probes}. Probe readout, however, does not share the asymmetry identified in Section~\ref{sec:finding1}. In particular, both Maze states encode global path information, although donor-$\zL$ patching produces a stronger transfer of the intermediate path state. Probe accuracy alone, therefore, does not identify the state or direction with the greatest causal relevance. ARC-AGI-2 provides a representation-level replication beyond Sudoku and Maze. Its probes recover grid geometry, color structure, object boundaries, and input--output transformation features from HRM activations. Because ARC-AGI-2 tasks are heterogeneous and may change grid shape, these results demonstrate that structured task information remains accessible in a broader grid-transformation setting.

\begin{figure*}[t]
\centering
\includegraphics[width=0.95\textwidth]{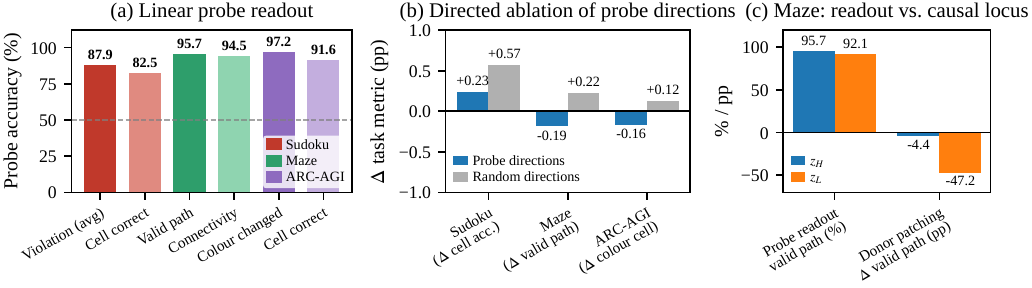}
\caption{\textbf{Probe readout does not imply causal relevance.}
\textbf{(a)} Task variables are linearly decodable from HRM states across Sudoku, Maze, and ARC-AGI-2.
\textbf{(b)} Directed ablation removes the corresponding probe directions and compares the behavioral effect with rank-matched random directions.
\textbf{(c)} Maze exhibits the mismatch most clearly: both $\zH$ and $\zL$ decode global path variables, whereas Section~\ref{sec:finding1} identifies donor-$\zL$ patching as the stronger intervention on the intermediate path state.}
\label{fig:readout_vs_causality}
\end{figure*}

\paragraph{Causal relevance of probe directions.} We perform directed ablations of the learned linear probe directions. For a unit probe direction $\hat{\mathbf{w}}$, we project the activation onto its orthogonal complement:
\begin{equation}
    z \leftarrow z-(z^\top\hat{\mathbf{w}})\hat{\mathbf{w}}.
\end{equation}
Each probe-direction ablation is compared with rank-matched random controls obtained by sampling directions from Gaussian distribution. We evaluate paired per-example changes and report bootstrap confidence intervals. From Figure~\ref{fig:readout_vs_causality}, probe-direction ablations on Sudoku and Maze produce small behavioral changes relative to full-state interventions and are comparable to random controls. The directions identified by the linear probes are therefore not causally privileged for HRM computation.

\paragraph{Summary.} These results answer RQ2: task variables are linearly decodable from $\zH$ and $\zL$, yet ablations exhibits substantially weaker effects than full-state interventions and comparable effects to random controls. Additional results in Appendix~\ref{app:mlp_probes_main} on nonlinear MLP probes support a similar conclusion. Thus, probe readout characterizes information available in the representation but does not by itself identify the directions on which HRM causally depends.

\begin{figure}[t!]
\centering
\includegraphics[width=.9\linewidth]{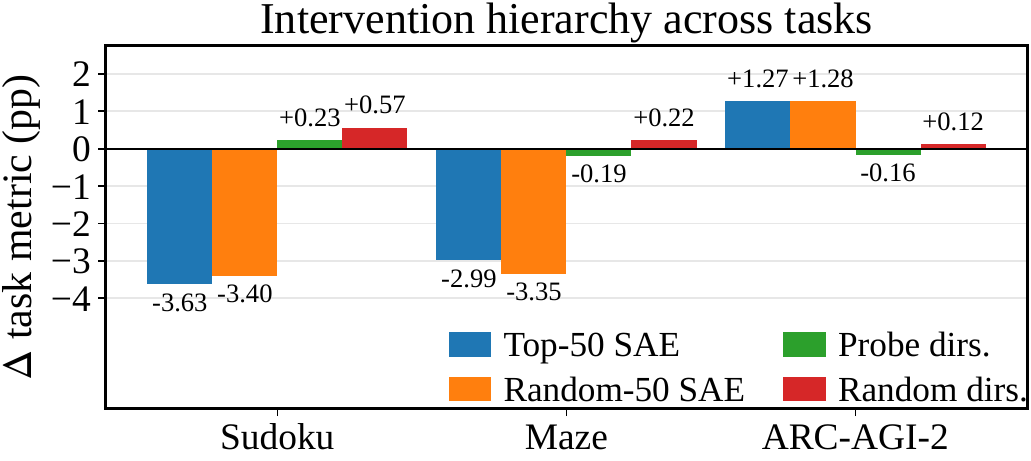}
\caption{Causal effects for a representative SAE configuration. Values are mean percentage-point changes relative to the no-intervention output; negative values indicate performance degradation. Sudoku reports $\Delta$cell accuracy, Maze reports $\Delta$valid path rate, and ARC-AGI-2 reports $\Delta$color-cell accuracy. Full SAE sweeps are reported in Appendix~\ref{app:sae_sweeps}.
}
\label{fig:causal_ladder}
\end{figure}

\section{Finding 3: HRM's Causal Computation Is Distributed within Latent Space}
\label{sec:finding3}
Section~\ref{sec:finding2} reveal that task variables are decodable from HRM recurrent latent states, while ablating the probe directions yields limited behavioral changes. This motivates us to study RQ3: whether a richer feature basis isolates a set of causally important features.

\paragraph{Setup.} We train sparse autoencoders (SAEs) on HRM recurrent activations to identify an over-complete, nonlinear dictionary of sparse activation features. Unless otherwise specified, we find a representative SAE hyperparameter settings via a grid search over dictionary size $D_{\mathrm{SAE}}\in\{1024,2048,4096\}$ and sparsity penalties
$\lambda\in\{0.001,0.003,0.01,0.03,0.1\}$ for Sudoku, Maze, and ARC-AGI-2. More details on hyperparameter search are presented in Appendix~\ref{app:sae_sweeps}. At intervention time, the activation is encoded, a selected subset of SAE coefficients is set to zero, and the resulting representation is decoded before inference continues. The top-$k$ features, ranked by activation magnitude, are compared with $k$ randomly selected features. Successful localization would predict larger behavioral effects under top-$k$ ablation than under random-$k$ ablation. Interventions are evaluated against the corresponding no-intervention output. We evaluate cell accuracy for Sudoku, valid path rate for Maze, and color-cell accuracy for ARC-AGI-2.


\paragraph{SAE feature analysis.} We have two key findings from Figure~\ref{fig:causal_ladder}. First, SAE-feature ablations are more behaviorally disruptive than probe-direction ablations in Section~\ref{sec:finding2}. On Sudoku and Maze, removing SAE features causes more task metric decrease than ablating probe-directions (about -3\% vs. near zero). Second, this causal effect is not concentrated in the highest-ranked SAE features. Across all tasks, top-ranked and randomly selected feature subsets achieves comparable performance. The learned dictionary, therefore, contains causally relevant structure, but the ranking criterion does not identify a privileged sparse subset. For $k = 50$, top and random subsets change the primary metric by $-3.63\%$ and $-3.40\%$ points on Sudoku with $p$-value $p = 0.43$, $-2.99\%$ and $-3.35\%$ on Maze with $p$-value $p = 0.18$, and $+1.27\%$ and $+1.28\%$ on ARC-AGI-2 with $p$-value $p = 0.94$. The same qualitative pattern is observed across different dictionary sizes and sparsity penalties in Appendix~\ref{app:sae_sweeps}.

Together with Sections~\ref{sec:finding1} and~\ref{sec:finding2}, these results demonstrate that full-state interventions produce the largest behavioral effects, SAE-feature ablations produce intermediate effects, and probe-direction ablations are weak or comparable to random directions. The larger effects of SAE ablation indicate that the recurrent state is causally consequential, but the equivalence between top-ranked and random SAE subsets indicates that this causal structure is not localized by the learned sparse basis.

\paragraph{One-step gradient approximation vs. BPTT.} The released HRM is trained using a one-step gradient approximation, motivated by deep-equilibrium-style implicit differentiation. Only the final recurrent step receives direct gradient tracking, while the recurrent carry is detached across adaptive computational time (ACT) steps. Prior work has questioned the suitability of this approximation when the recurrent dynamics are not guaranteed to converge to the assumed fixed point \citep{jolicoeur2025tiny}.

Thus, we ask whether the training procedure may contribute to the distributed SAE representation. In particular, localized sparse features may fail to emerge if gradients do not account for intermediate recurrent computation. Under this hypothesis, the top-versus-random equivalence in Figure~\ref{fig:causal_ladder} would reflect the one-step approximation rather than the learned computational strategy.

This hypothesis is evaluated through a controlled comparison between the vanilla HRM using one-time gradient approximation and the counterpart using backpropagation-through-time (BPTT). Both models are trained from scratch using matched configurations, data, and random seeds. The SAE intervention analysis is then repeated on a shared set of 200 Sudoku puzzles.

\begin{table}[t]
\centering
\caption{BPTT training-regime check. Values are mean changes in cell accuracy under SAE-feature ablation over 200 shared puzzles. Top-50 and random-50 ablations are statistically indistinguishable in both regimes (paired $t$-tests: stock, $p=0.39$, $d=-0.06$; BPTT, $p=0.93$, $d=-0.01$).}
\label{tab:bptt_check}
\vskip 0.1in
\small
\begin{tabular}{@{}lcc@{}}
\toprule
\textbf{Model} & \textbf{Top-50 SAE} & \textbf{Random-50 SAE} \\
\midrule
Stock HRM & $-1.59$  & $-1.37$ \\
BPTT HRM  & $-29.64$ & $-29.63$ \\
\bottomrule
\end{tabular}
\end{table}

Table~\ref{tab:bptt_check} distinguishes the magnitude of the SAE intervention effect from its concentration among the top-ranked features. The intervention magnitude depends strongly on the training regime. Under BPTT, top-50 feature ablation reduces cell accuracy by $29.64\%$, compared with $1.59\%$ for the vanilla model. SAE-feature removal, therefore, has a substantially larger impact in the BPTT-trained model. However, top-$k$ ranked features are not causally more important than $k$ randomly selected features either when trained with BPTT. In the vanilla model, top-50 and random-50 ablations reduce accuracy by $1.59\%$ and $1.37\%$, respectively. Under BPTT, the corresponding reductions are $29.64\%$ and $29.63\%$. Neither regime exhibits a statistically significant difference between top-ranked and randomly selected feature subsets. BPTT, therefore, increases the causal effect of SAE feature ablation while still being indistinguishable from random feature ablation. The statistically insignificant top-versus-random differences  for both vanilla HRM and BPTT-trained counterpart indicate that one-step gradient approximation is not the root cause of such distributed computation across latent states.

Because the vanilla and BPTT-trained models are trained independently, their absolute intervention magnitudes are not directly comparable with the cross-task results from Figure~\ref{fig:causal_ladder}. We provide full training, SAE-sweep, and evaluation details in Appendix~\ref{app:bptt}.

\paragraph{Summary.} The recurrent state in HRM contains causally relevant structure that is not localized by the feature bases examined in this study. Linear probes recover constraint-aware directions, but these directions have weak causal effects, as established in Section~\ref{sec:finding2}. SAE features show larger behavioral effects, yet top-ranked feature subsets are no more important than randomly selected subsets. This pattern is consistent across Sudoku, Maze, and ARC-AGI-2, remains stable across SAE configurations, and persists under BPTT training. Although BPTT substantially increases the effect of SAE-feature removal, it does not concentrate causal relevance in the highest-ranked features. These results indicate that HRM does not simply rely on top-ranked features from constraint-aware probes nor SAEs, but rather distributing the computation and reasoning across the recurrent latent states.

\section{Discussion and Conclusion}~\label{sec:discussion}
From our findings, we characterize HRM as performing \emph{constraint-aware iterative refinement on a puzzle specific solution state that is allocated flexibly across its hierarchy and distributed within it}. 

\paragraph{Hierarchy as an interpretability feature.}
Our compare HRM against four Transformer-based baselines and observe: a flat recurrent transformer matches HRM on Sudoku-Extreme. The hierarchical split is not what makes HRM strong; rather, it provides two individually intervenable latent states that are causally separable (Sections~\ref{sec:finding1}). Because the roles of the modules are task-specific, the value of hierarchy shows itself in this separability. Hierarchy thus can serve as an \emph{inductive bias for interpretability}.

\paragraph{Linearly decodable probe readout $\neq$ causally relevant information.}
From Sections~\ref{sec:finding2} and \ref{sec:finding3}, we observe that easily decodable directions in $\zH$ and $\zL$ have no measurable causal effect when ablated, and an over-complete SAE recovers no mono-semantic features for the symbolic targets tested. Thus, studies based on probes risk overstating feature relevance.

\paragraph{Representations might be more distributed in HRM than in feed-forward LLMs.}
The BPTT experiments rule out one of the most concrete candidate explanations: the one-step gradient approximation during HRM training. Two hypotheses remain open now. First, a single $512$-dimensional state reused across all positions and ACT steps must simultaneously encode static input structure, intermediate constraint information, and the evolving solution; this would likely favor dense, superposed information over mono-semantic features \citep{bricken2023monosemanticity, templeton2024scaling}. Second, magnitude-ranked SAE features may be the wrong unit for recurrent states, BPTT experiments suggest that the dictionary captures a causally relevant subspace rather than discrete features.

\paragraph{Conclusion.}
We present an interpretability-centric analysis to understand the working mechanism of HRM across Sudoku, Maze, and ARC-AGI-2. From the analysis, we have three key findings: (i) HRM iteratively refines a puzzle-specific solution state, and the functional contribution of each component across its hierarchy is task-dependent (Section~\ref{sec:finding1}). (ii) Task specific features are highly decodable from both high- and low-level activations, yet the corresponding probe directions do not display any causal relevance (Section~\ref{sec:finding2}). While this readout-causality dissociation is well documented in feed-forward models \citep{hewitt2019designing, elazar2021amnesic}, our contribution establishes the result for recursive, latent-reasoning models, where probes must be validated across a continually refined state. (iii) SAE features discover more causally important features than probe directions, but top-ranked features are no more causally important than random features, a pattern that persists across hyperparameter settings, tasks, and under full BPTT training (Section~\ref{sec:finding3}). Together, these findings characterize HRM as a flexible two stream, iterative refinement model whose causally relevant state is distributed across its latent representations. The findings also highlight the significance and requirement of mechanistic interpretability techniques better suited for latent-space, recursive reasoning models. 

\paragraph{Limitations.}
Three limitations bound our claims: (i) ARC-AGI-2 serves only as a representational-level replication, since its heterogeneous tasks make module level causal claims hard to interpret, (ii) BPTT check covers Sudoku only and compares from-scratch models whose absolute effect sizes are not comparable to the released checkpoint, (iii)the project focuses on the $\sim$27M parameter HRM model, whether other latent-reasoning models of different sizes have different internal organizations remains open. 

\paragraph{Future Work.}
Natural future work would include (i) testing alternative feature bases and rankings for recurrent activations such as attribution-ranked SAE features, transcoders, or cross step dictionaries; (ii) training objectives that explicitly promote sparse, localizable features, and (iii) extending the analysis to other recursive latent reasoners like TRMs \citep{jolicoeur2025tiny}.

\bibliography{aaai2027}

\clearpage

\appendix
\section{Additional Experimental Details}
\label{app:details}


\subsection{Full HRM Architecture and Training Details}
\label{app:hrm}

We expand the Hierarchical Reasoning Model (HRM) overview in Section~\ref{sec:background} and discuss the training procedure of HRM~\citep{wang2025hrm}.


\paragraph{Components.}
HRM comprises (1) an input token embedding network $f_\mathrm{I}: \mathcal{V}^{S}\rightarrow \mathbb{R}^{S \times d}$, where $S$ denotes the number of input positions and $d$ denotes the hidden dimension, (2) separate high- and low-level recurrent modules $f_H$ and $f_L$, and (3) an output head $f_\mathrm{O}: \mathbb{R}^{S \times d} \rightarrow \mathbb{R}^{S \times |\mathcal{V}|}$. The recurrent modules are $4$ layers of encoder-only Transformer~\citep{vaswani2017attention} with bi-directional self-attention, SwiGLU feed-forward blocks~\citep{shazeer2020glu}, rotary position embeddings (RoPE)~\citep{su2024roformer}, and post-norm RMS Norm~\citep{zhang2019root}. The two modules share the same architecture (hidden dimension $d = 512$ and $8$ attention heads) with different initialization. The model contains ${\sim}27$ million parameters in total.

\paragraph{Recurrent update.}
Let $\mathbf{x} \in \mathcal{V}^{S}$ be the input sequence of $S$ tokens, where $\mathcal{V}$ is the vocabulary. HRM first extracts the token embedding via the input token embedding network
\begin{equation}
    \tilde{\mathbf{x}} = f_{\mathrm I}\left(\mathbf{x}\right)
\end{equation}
It then maintains the high and low level carry states $\zH \in \mathbb{R}^{S \times d}$ and $\zL \in \mathbb{R}^{S \times d}$, respectively. A forward pass (i.e., a segment) of HRM consists of $N$ high-level updates, each with $T$ low-level updates. Specifically, 
\begin{align}
    \zL &\leftarrow f_{\rm L}\left(\zL,\; \zH + \tilde{\mathbf{x}}\right), \label{eq:zL-update} \\
    \zH &\leftarrow f_{\rm H}\left(\zH,\; \zL\right), \label{eq:zH-update}
\end{align}
where the second argument is additively injected into the running state:
$\zH+\tilde{\mathbf{x}}$ is injected into $\zL$ in the low-level update,
whereas $\zL$ is injected into $\zH$ in the high-level update. We follow the setting provided by \citet{wang2025hrm} and set $N = T = 2$, meaning that each segment performs $2$ high-level and $4$ ($N \times T = 4$) low-level updates. After this, $f_{\rm O}$ maps $\zH$ to per-position logits. The resulting $\left(\zL, \zH\right)$ is detached from the computation graph and propagated to the next segment.

\paragraph{Adaptive halting.}
After each segment, an adaptive halting head for adaptive computation time (ACT)~\citep{graves2016act} predicts whether to halt or continue the training process, up to $M_{\max} = 16$ segments. The pair $\left(\zH, \zL\right)$ propagated between segments is the \textit{carry}; it is reset once the current input is marked as halted or when the input $\mathbf{x}$ changes.

\paragraph{Training procedure.}
HRM is trained with deep supervision. Specifically, a prediction loss $\mathcal{L}$ is computed after each segment, and the final loss function is $\sum_{t = 0}^{M_{\max} - 1}\mathcal{L}\left(\hat{\mathbf{y}}^{\left(t\right)},\mathbf{y}\right)$. In this way, each segment learns to produce a usable prediction without backpropagating through earlier segments. One-step gradient approximation is applied in each segment. That is, only the final $\zL$ and $\zH$ updates within a segment are differentiated for optimization. The adaptive halting head is trained using Q-learning, so the predicted halt value approximates the expected solution quality of the current segment. We use the publicly released checkpoints \footnote{\url{https://huggingface.co/sapientinc}} for our experiments, except for the BPTT comparison in Section~\ref{sec:finding3} and Appendix~\ref{app:bptt}, which trains HRM from scratch.


\subsection{Tasks, Datasets, Evaluation Details}
\label{app:dataset}
Unless otherwise specified, we disable learned early stopping and execute all $16$ segments for mechanistic understanding, indexed by step $t = \left\{0, \ldots, 15\right\}$.

\paragraph{Sudoku-Extreme.}
The Sudoku-Extreme dataset is constructed from publicly available Sudoku corpora. It is also filtered to keep only puzzles whose human-solver rating is ``extreme''. These puzzles require multi-step logical deductions and long sequences of constraint propagation. We follow the dataset generation procedure by \citet{wang2025hrm} and build a $1{,}001{,}000$-puzzle training corpus by drawing a uniform random sample of $1{,}000$ base puzzles and adding $1{,}000$ structure-preserving augmentations per puzzle. Evaluations use a deterministic subset of the first $500$ puzzles retrieved from the test data loader.


\paragraph{Maze-Hard.} We use the publicly released dataset\footnote{\url{https://huggingface.co/datasets/sapientinc/maze-30x30-hard-1k}} and the corresponding released checkpoint without retraining. The puzzles are $30\times 30$ grids encoding blocked, free, start, and goal cells, with a valid path from start to goal as a label. Interventions and probes use $1{,}000$ held-out test mazes drawn deterministically from the test loader with \texttt{seed=42}. Probes are trained at steps $\{0,1,2,4,8,15\}$ on $\zH$ and $\zL$ over $5$ seeds. SAE ablation experiments use $300$ test mazes with start-goal path validity as the primary metric.


\paragraph{ARC-AGI-2.}
ARC is used as a representation-level replication rather than primary causal evidence. We rebuild the ARC-AGI-2 augmented dataset with \texttt{seed=42} and \texttt{number of augmentations = 1000}. These augmentations include operations like rotate/reflect a grid, and permute colors. Starting from a publicly released checkpoint (\texttt{arc2-adapted-evalonly}, step $7{,}391$, where step refers to the number of parameter updates), we fine-tune a model using the rebuilt dataset. Probes are trained with activations from $\zH$ and $\zL$ across steps $\left\{0, 1, 2, 4, 8, 15\right\}$ and 5 seeds $\{0,1,2,3,4\}$. SAE experiments use $150$  puzzles with color-cell accuracy as the primary metric.

\subsection{Datasets and Baselines for Finding 1}
\label{app:baseline_recurrence}

\begin{table}[!htbp]
\centering
\caption{Architectures compared in this paper. Recur. denotes refinement-step recurrence; Hier. denotes hierarchical structure; Steps denotes max encoder updates per input.}
\label{tab:architectures}
\vskip 0.1in
\small
\begin{tabular}{@{}lcccr@{}}
\toprule
\textbf{Model} & \textbf{Recur.} & \textbf{Hier.} & \textbf{Steps} & \textbf{Params} \\
\midrule
HRM                  & \checkmark & \checkmark & 16 & 27M \\
Recurrent Trans.     & \checkmark &            & 16 & 27M \\
Univ.\ Trans.        & \checkmark &            & 16 & 28M \\
Plain Trans.         &            &            & 1  & 27M \\
GRU                  &            &            & 1  & 25M \\
\bottomrule
\end{tabular}
\end{table}

We compare HRM against four baselines in Tables~\ref{tab:architectures}. All baselines use the same input embedding $f_{\rm I}$ and output head $f_{\rm O}$ as HRM and are trained on the corpus described in Appendix~\ref{app:dataset}. These ensure that the only difference lies in the reasoning module used. Recurrent Transformer updates a single hidden state with an $8$-layer post-norm Transformer, and the Universal Transformer is its weight shared variant~\citep{dehghani2019universal}. The Plain transformer ($8$-layer feedforward)~\citep{vaswani2017attention} and GRU~\citep{cho2014gru} are single-pass models without any recurrence. For all baselines, the number of trainable parameters range from 25M to 28M. We compare the performance of all models on the Sudoku-Extreme task. 




\begin{table}[t]
\centering
\caption{Baseline metrics across training checkpoints on 500 held-out Sudoku puzzles. Recurrent models are evaluated at segment~15; non-recurrent models at their single forward pass. Bold marks the best checkpoint of each model.}
\label{tab:baseline_checkpoints}
\vskip 0.1in
\small
\begin{tabular}{@{}lccc@{}}
\toprule
\textbf{Checkpoint} & \textbf{Cell\%} & \textbf{Puzzle\%} & \textbf{Hamming} \\
\midrule
Recurrent Trans.\ epoch 2 & 76.2 & 34.4 & 19.3 \\
Recurrent Trans.\ epoch 4 & \textbf{81.4} & \textbf{51.4} & \textbf{15.1} \\
Recurrent Trans.\ epoch 5 & 74.3 & 32.6 & 20.8 \\
Recurrent Trans.\ epoch 6 & 71.7 & 25.4 & 22.9 \\
\midrule
Univ.\ Trans.\ epoch 3 & 77.0 & 35.4 & 18.6 \\
Univ.\ Trans.\ epoch 4 & \textbf{80.7} & \textbf{48.4} & \textbf{15.6} \\
Univ.\ Trans.\ epoch 6 & 66.3 &  9.6 & 27.3 \\
Univ.\ Trans.\ epoch 8 & 66.6 &  9.8 & 27.0 \\
\midrule
Plain Trans.\ epoch 2 & 60.9 & 0.2 & 31.7 \\
Plain Trans.\ epoch 3 & \textbf{61.9} & \textbf{0.2} & \textbf{30.9} \\
Plain Trans.\ epoch 4 & 61.0 & 0.4 & 31.6 \\
Plain Trans.\ epoch 5 & 61.0 & 0.4 & 31.6 \\
Plain Trans.\ epoch 6 & 60.1 & 0.4 & 32.3 \\
Plain Trans.\ epoch 7 & 60.8 & 0.0 & 31.7 \\
Plain Trans.\ epoch 8 & 60.7 & 0.0 & 31.8 \\
\midrule
GRU epoch 2 & 48.0 & 0.0 & 42.1 \\
GRU epoch 3 & 49.2 & 0.0 & 41.1 \\
GRU epoch 4 & 49.4 & 0.0 & 41.0 \\
GRU epoch 5 & 49.5 & 0.0 & 40.9 \\
GRU epoch 6 & 49.5 & 0.0 & 40.9 \\
GRU epoch 7 & 49.6 & 0.0 & 40.9 \\
GRU epoch 8 & \textbf{49.6} & \textbf{0.0} & \textbf{40.8} \\
\bottomrule
\end{tabular}
\end{table}

\subsection{Training Details for Finding 1}
\label{app:baselines}
\begin{figure}[!htbp]
\centering
\includegraphics[width=\columnwidth]{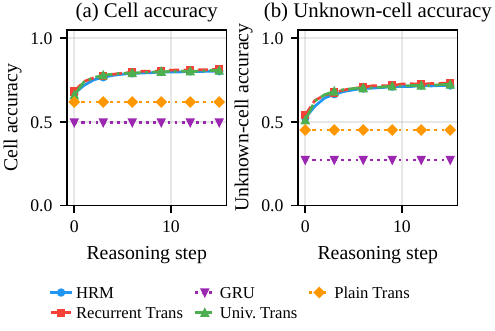}
\caption{Cell accuracy across segments for all five architectures on the held-out 500-puzzle test set. The recurrent models (HRM, Recurrent Transformer, Universal Transformer) improve monotonically; the non-recurrent baselines (Plain Transformer, GRU) are flat single-pass.}
\label{fig:accuracy_curves_5model}
\end{figure}

\begin{figure}[!htbp]
\centering
\includegraphics[width=\columnwidth]{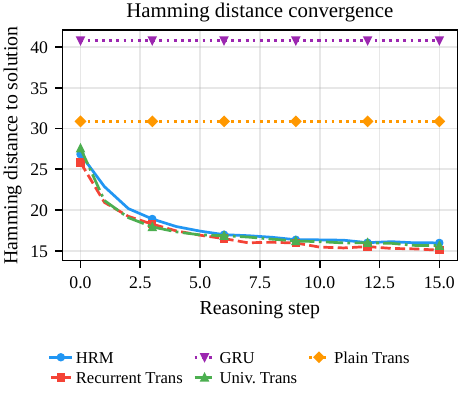}
\caption{Hamming distance to the ground-truth solution across segments for the same five architectures.}
\label{fig:hamming_convergence}
\end{figure}

All baselines follow the training procedure of HRM on the same Sudoku-Extreme corpus (Appendix ~\ref{app:dataset}). We follow the same optimizer/learning rate/batch size schedule as described by \citet{wang2025hrm}. All models are evaluated on the same 500-puzzle test set under a maximum of 16 segments (i.e., recurrence steps). We report the test performance of all trained models in Table~\ref{tab:baseline_checkpoints}. Cell accuracy and Hamming distance across all recurrence steps for all models are presented in Figures ~\ref{fig:accuracy_curves_5model} and~\ref{fig:hamming_convergence}. Note that for the plain Transformer and GRU, we use straight lines to indicate their performances, as they are not recurrent. From these results, the recurrent architectures significantly outperform non-recurrent baselines. Recurrent and Universal transformers are able to reach a puzzle accuracy of $51.4\%$ and $48.4\%$ respectively, while the Plain Transformer and GRU solve (almost) no complete puzzles (Table~\ref{tab:baseline_checkpoints}). The step-wise curves for recurrent baselines also show improvements in accuracy, while non-recurrent variants remain constant throughout. These results indicate that strong performance on tasks like Sudoku is driven by recurrence rather than parameter count. 




\subsection{Full Intervention Results for Finding 1}
\label{app:interventions}

Here, we provide additional results to Section~\ref{sec:finding1}.
All interventions share a common protocol: activations are frozen,
ablated, replaced, or transplanted during the forward pass. We compare
each intervention with a clean, no-intervention rollout of the same
target puzzle. We report the percentage-point change
\begin{equation}
    \Delta M = M_{\rm intervention} - M_{\rm clean},
\end{equation}
where $M$ is the task-specific measurement: cell accuracy for Sudoku
and valid start--goal path rate for Maze, unless otherwise stated.
Sample sizes and clean baselines are reported in each table caption.


\paragraph{Ablation.}
Table~\ref{tab:ablation_full} reports the full zero-ablation results
for Sudoku and Maze.

For Sudoku, single-step $\zH$ ablation reduces cell accuracy more near the final step. In contrast, single-step $\zL$ ablation has negligible effects, with no selected step exceeding a $1.0$ pp reduction in cell accuracy. This is consistent with our finding in Section~\ref{sec:finding1}, where $\zH$ carries the solution state for Sudoku and $\zL$ acts as a working buffer. As the solution state progressively refines the final solution, ablating $\zH$ near the
final step causes increasingly severe errors. 

For Maze, token accuracy changes by less than $1$ pp under all-step ablation because it is dominated by static layout cells. Path validity, however, decreases substantially, confirming that endpoint token accuracy is a weak diagnosis of computation in Maze. Per-step ablation is comparatively small until the final step, when ablating $\zH$ reduces valid-path rate by $38.5$ pp. Single-step $\zL$ ablation never exceeds a $4.5$ pp reduction.

\begin{table*}[!t]
\centering
\caption{
Combined zero-ablation results. Panel A reports all-step ablation under the previously reported task metrics. Panel B reports the task-specific primary metric at selected steps $k$: cell accuracy for Sudoku ($1{,}000$ puzzles; clean baseline $81.8\%$) and valid start--goal path rate for Maze ($200$ puzzles; clean baseline $93.5\%$). All values are percentage-point changes from the corresponding clean rollout. ``All'' zeroes the indicated stream at every recurrent step.}

\label{tab:ablation_full}
\vskip 0.1in
\small

\textbf{Panel A: All-step ablation}
\vskip 0.05in
{\setlength{\tabcolsep}{5pt}
\begin{tabular}{@{}lrrrrr@{}}
\toprule
& \multicolumn{1}{c}{\textbf{Sudoku}} &
  \multicolumn{4}{c}{\textbf{Maze}} \\
\cmidrule(lr){2-2} \cmidrule(l){3-6}
\textbf{State} &
\textbf{Cell acc.} &
\textbf{Token acc.} &
\textbf{Path-cell acc.} &
\textbf{Valid S--G} &
\textbf{Connected} \\
\midrule
$\zH$ & $-19.3$ & $-0.8$ & $-2.7$ & $-32.5$ & $-26.0$ \\
$\zL$ & $-19.3$ & $-0.4$ & $-1.4$ & $-8.5$  & $-7.0$  \\
\bottomrule
\end{tabular}
}

\vskip 0.12in
\textbf{Panel B: All-step and single-step ablation}
\vskip 0.05in
{\setlength{\tabcolsep}{4pt}
\begin{tabular}{@{}llrrrrrrrr@{}}
\toprule
\textbf{Task} & \textbf{State} &
\textbf{All} & $\mathbf{1}$ & $\mathbf{2}$ & $\mathbf{4}$ &
$\mathbf{8}$ & $\mathbf{12}$ & $\mathbf{14}$ & $\mathbf{15}$ \\
\midrule
Sudoku & $\zH$ & $-19.3$ & $-1.8$ & $-3.3$ & $-4.3$ &
                  $-7.2$ & $-13.0$ & $-19.4$ & $-26.0$ \\
Sudoku & $\zL$ & $-19.3$ & $-0.3$ & $-0.7$ & $-0.2$ &
                  $-0.1$ & $-0.3$ & $-0.5$ & $-1.0$ \\
\midrule
Maze   & $\zH$ & $-32.5$ & $-2.0$ & $-2.0$ & $-2.0$ &
                  $-2.5$ & $-2.5$ & $-6.5$ & $-38.5$ \\
Maze   & $\zL$ & $-8.5$  & $-1.5$ & $-1.0$ & $-1.0$ &
                  $-0.5$ & $-0.5$ & $-0.5$ & $-4.5$ \\
\bottomrule
\end{tabular}
}
\end{table*}


\paragraph{Freezing.}
Table~\ref{tab:freezing_full} reports the effect of freezing each latent stream from step $k$ onward.

For Sudoku, freezing $\zH$ at earlier steps is harmful, while the loss decreases as $k$ approaches the final step. Freezing $\zL$ has a negligible effect at every selected step. This is consistent with a solution state in $\zH$ that progressively accumulates information: freezing $\zH$ at later steps preserves more of the information needed to derive the correct Sudoku solution.

For Maze, freezing either stream from any step changes token accuracy by at most $\pm 0.02$ pp. The valid-path results in Table~\ref{tab:freezing_full} are similarly small. Unlike Sudoku, Maze mostly reaches a final solution during the first recurrent updates, leaving comparatively little subsequent refinement for freezing to affect.

\begin{table*}[!t]
\centering
\caption{
Combined freezing results. Each stream is frozen from step $k$ onward. Sudoku reports $\Delta$cell accuracy ($1{,}000$ puzzles; clean baseline $81.8\%$), while Maze reports $\Delta$valid start--goal path rate ($1{,}000$ puzzles; clean baseline $92.4\%$). Values are percentage points.}
\label{tab:freezing_full}
\vskip 0.1in
\small
{\setlength{\tabcolsep}{5pt}
\begin{tabular}{@{}llrrrrrrr@{}}
\toprule
\textbf{Task} & \textbf{State} &
$\mathbf{1}$ & $\mathbf{2}$ & $\mathbf{4}$ & $\mathbf{8}$ &
$\mathbf{12}$ & $\mathbf{14}$ & $\mathbf{15}$ \\
\midrule
Sudoku & $\zH$ & $-6.2$ & $-3.9$ & $-2.0$ & $-0.4$ &
                  $-0.2$ & $0.0$ & $0.0$ \\
Sudoku & $\zL$ & $+0.2$ & $-0.2$ & $+0.2$ & $+0.2$ &
                  $-0.1$ & $0.0$ & $0.0$ \\
\midrule
Maze   & $\zH$ & $-0.5$ & $-0.4$ & $+0.7$ & $+0.9$ &
                  $+0.6$ & $0.0$ & $0.0$ \\
Maze   & $\zL$ & $+0.5$ & $0.0$ & $+0.1$ & $0.0$ &
                  $+0.3$ & $0.0$ & $0.0$ \\
\bottomrule
\end{tabular}
}
\end{table*}


\paragraph{Within-instance temporal transplant.}
Within-instance temporal transplant replaces the $\zH$ state at one recipient step with the state from a different donor step of the same puzzle. Selected results are reported in Table~\ref{tab:temporal_transplant_full}.

For Sudoku, future-to-past transplants slightly improve accuracy, while the inverse direction hurts. Transplants into step~2 yield gains of $+0.6$ to $+0.8$ pp across later donor steps, while transplants from step~10 into steps~13--15 yield losses of $-0.2$ to $-0.6$ pp. The asymmetry confirms monotonic refinement: later states are improvements over earlier ones.

For Maze, all within-instance temporal transplants change token accuracy by less than $0.07$ pp. The corresponding valid-path changes also remain small, as shown in Table~\ref{tab:temporal_transplant_full}. This is consistent with the early saturation described in Appendix~\ref{app:traj_pca}.

\begin{table*}[!t]
\centering
\caption{
Selected within-instance $\zH$ temporal transplants. Column labels give
the donor and recipient steps as donor${\to}$recipient. Sudoku reports
$\Delta$cell accuracy and Maze reports $\Delta$valid start--goal path
rate. Both experiments use $1{,}000$ puzzles, with clean baselines of
$81.8\%$ for Sudoku and $92.4\%$ for Maze. Values are percentage points.
}
\label{tab:temporal_transplant_full}
\vskip 0.1in
\small
{\setlength{\tabcolsep}{6pt}
\begin{tabular}{@{}lrrrrrr@{}}
\toprule
\textbf{Task} &
$\mathbf{9{\to}2}$ &
$\mathbf{12{\to}2}$ &
$\mathbf{15{\to}2}$ &
$\mathbf{10{\to}13}$ &
$\mathbf{10{\to}14}$ &
$\mathbf{10{\to}15}$ \\
\midrule
Sudoku & $+0.8$ & $+0.7$ & $+0.6$ & $-0.2$ & $-0.5$ & $-0.6$ \\
Maze   & $+0.3$ & $+0.8$ & $+0.3$ & $+0.2$ & $+0.6$ & $+0.6$ \\
\bottomrule
\end{tabular}
}
\end{table*}


\paragraph{Cross-puzzle patching.}
Cross-puzzle patching replaces the target puzzle's latent state at step $k$ with the corresponding state from a different donor puzzle. Table~\ref{tab:patching_full} reports the results on the common step grid.

\begin{figure}[t]
\centering
\includegraphics[width=0.90\columnwidth]
{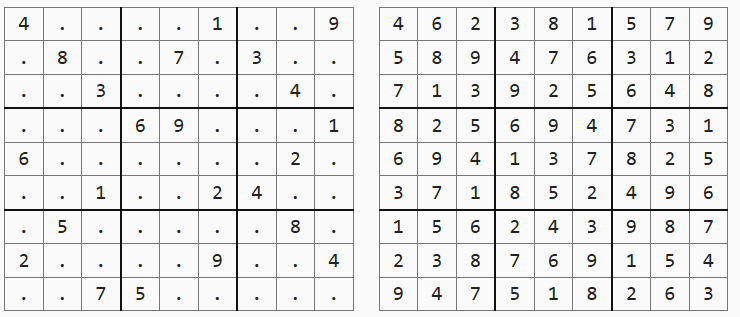}
\caption{Recipient puzzle used in
Figure~\ref{fig:patching_grids}: given clues (left) and ground-truth
solution (right).}
\label{fig:patching_target_app}
\end{figure}

\begin{figure}[t]
\centering
\includegraphics[width=0.90\columnwidth]
{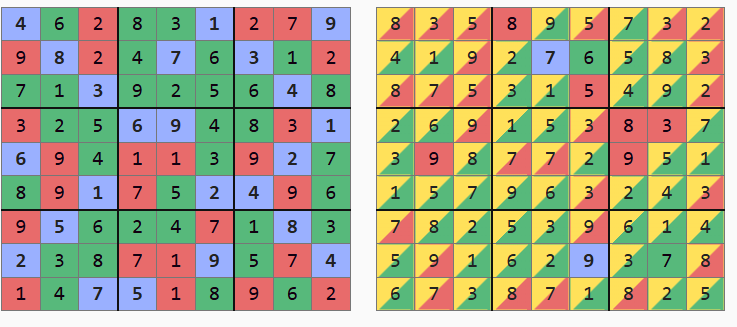}
\caption{Sudoku cross-puzzle patching example: baseline prediction on
the target puzzle (left) versus prediction after donor-$\zH$ patching
(right). The patched output is rewritten toward the constraints of the
donor puzzle. Blue denotes a given cell, green denotes a
constraint-satisfying prediction, red denotes a constraint-violating
prediction, and yellow denotes a prediction changed from the previous
recurrent step. A combination of yellow and red/green denotes a cell
that changed value and is constraint-violating/satisfying.}
\label{fig:patching_grids}
\end{figure}

For Sudoku, donor-$\zH$ patching rewrites the output toward the constraints of the donor at every tested stage, whereas donor-$\zL$ patching does not exhibit this behavior. The effect is present at the earliest patchable step rather than emerging only at readout. Figures~\ref{fig:patching_grids} and~\ref{fig:patching_target_app} show a representative case.

For Maze, donor-$\zL$ patching decreases valid-path rate by $43$--$56$ pp at every tested step. Donor-$\zH$ patching has little effect before the final step, where valid-path rate decreases by $26$pp. Since single-step $\zL$ ablation never exceeds a $4.5$pp reduction, the causal role of $\zL$ is highlighted more strongly through patching than through zero-ablation.

\begin{table*}[!t]
\centering
\caption{
Combined cross-puzzle patching results on the common step grid.
Both tasks are evaluated over $250$ donor--target pairs.
Sudoku reports $\Delta$cell accuracy (clean target baseline $78.3\%$).
Maze reports $\Delta$valid start--goal path rate under full-grid
patching (clean target baseline $92.0\%$).
Values are percentage points.
}
\label{tab:patching_full}
\vskip 0.1in
\small
{\setlength{\tabcolsep}{5pt}
\begin{tabular}{@{}llrrrrrrr@{}}
\toprule
\textbf{Task} & \textbf{Donor state} &
$\mathbf{1}$ & $\mathbf{2}$ & $\mathbf{4}$ & $\mathbf{8}$ &
$\mathbf{12}$ & $\mathbf{14}$ & $\mathbf{15}$ \\
\midrule
Sudoku & $\zH$ & $-55.8$ & $-58.8$ & $-60.6$ & $-61.7$ &
                  $-62.7$ & $-65.4$ & $-66.7$ \\
Sudoku & $\zL$ & $-0.3$  & $-1.5$  & $-2.3$  & $-0.2$ &
                  $-0.0$  & $+0.3$  & $-0.1$ \\
\midrule
Maze   & $\zH$ & $-1.6$  & $+0.4$  & $-0.8$  & $+0.8$ &
                  $+0.4$  & $-4.8$  & $-34.0$ \\
Maze   & $\zL$ & $-40.8$ & $-43.6$ & $-43.2$ & $-44.0$ &
                  $-44.4$ & $-42.8$ & $-56.0$ \\
\bottomrule
\end{tabular}
}
\end{table*}


\subsection{Latent Trajectory and PCA Analysis of Activations}
\label{app:traj_pca}

We study whether the geometry of recurrent trajectories corroborates the causal results. Inspired by the PCA visualization by \citet{ren2026guessing}, we conduct a population-level analysis of solved and failed Sudoku and Maze trajectories.

\begin{figure}[t]
\centering
\includegraphics[width=\columnwidth]{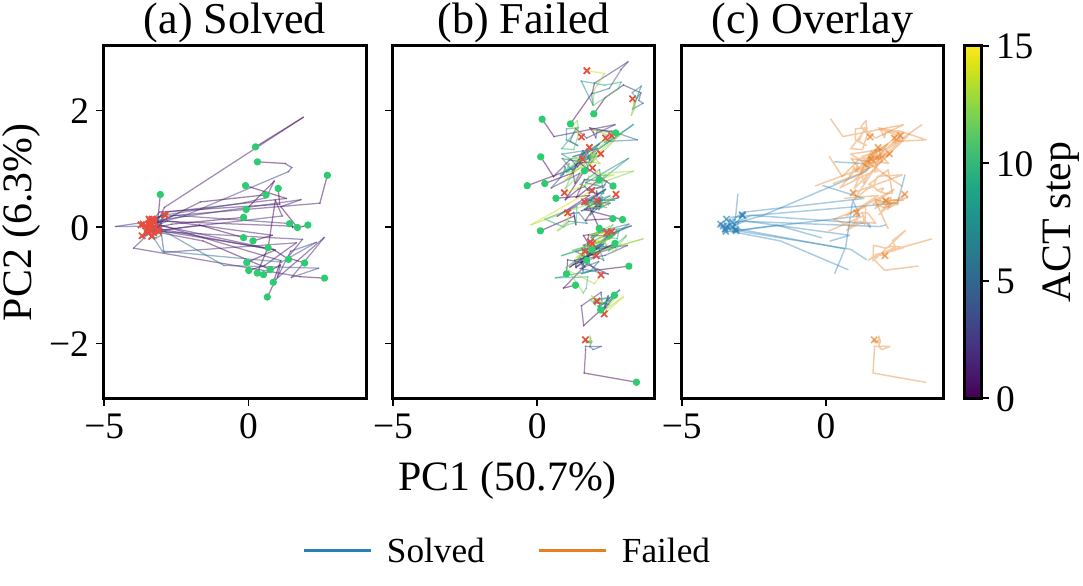}
\caption{PCA trajectories of $\zH$ across 16 segments for 200 Sudoku puzzles. \textbf{(a)}~Solved puzzles converge to a compact cluster. \textbf{(b)}~Failed puzzles remain broadly dispersed, including at their endpoints. \textbf{(c)}~Overlay of solved and failed trajectories, which occupy distinct regions of principal-component space. Green dots denote step~0 and red crosses denote step~15.}
\label{fig:zh_trajectories}
\end{figure}

\begin{figure}[t]
\centering
\includegraphics[width=\columnwidth]{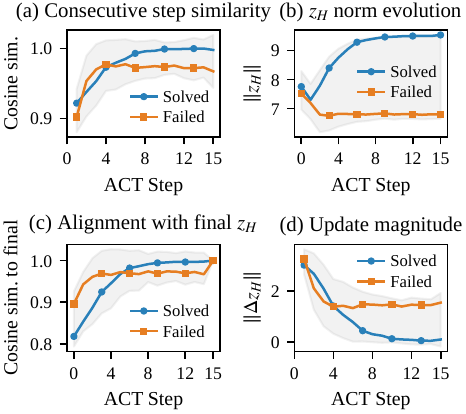}
\caption{Population-level metrics for $\zH$ across 16 segments, separated into solved and failed Sudoku puzzles: \textbf{(a)}~consecutive-state cosine similarity, $\cos(\zH^{(t)},\zH^{(t-1)})$; \textbf{(b)}~state norm, $|\zH^{(t)}|$; \textbf{(c)}~alignment with the final state; and \textbf{(d)}~update magnitude, $|\Delta\zH^{(t)}|$.}
\label{fig:zh_metrics}
\end{figure}

\paragraph{Sudoku: convergence versus wandering.}
We collect mean-pooled $\zH$ activations at each segment for 200 puzzles, comprising 93 solved and 107 failed examples, and project them onto the first two principal components, which explain $57\%$ of the variance. 
Solved and failed trajectories differ in both position and dynamics (Figure~\ref{fig:zh_trajectories} and ~\ref{fig:zh_metrics}). Solved puzzles converge to a compact cluster, with consecutive-state cosine similarity above $0.999$, and update magnitude $|\Delta\zH|$ decreasing from approximately $3.0$ to $0.09$. Failed puzzles remain dispersed and continue to oscillate. These results provide a quantitative counterpart to the ``spurious attractor'' phenomenon described qualitatively by \citet{ren2026guessing} and establish convergence behavior as a step-resolved signature of successful computation, complementing the causal evidence that $\zH$ carries the progressively refined solution.

\paragraph{Maze: low-dimensional trajectories and early saturation.}

\begin{figure}[t]
\centering
\includegraphics[width=\columnwidth]{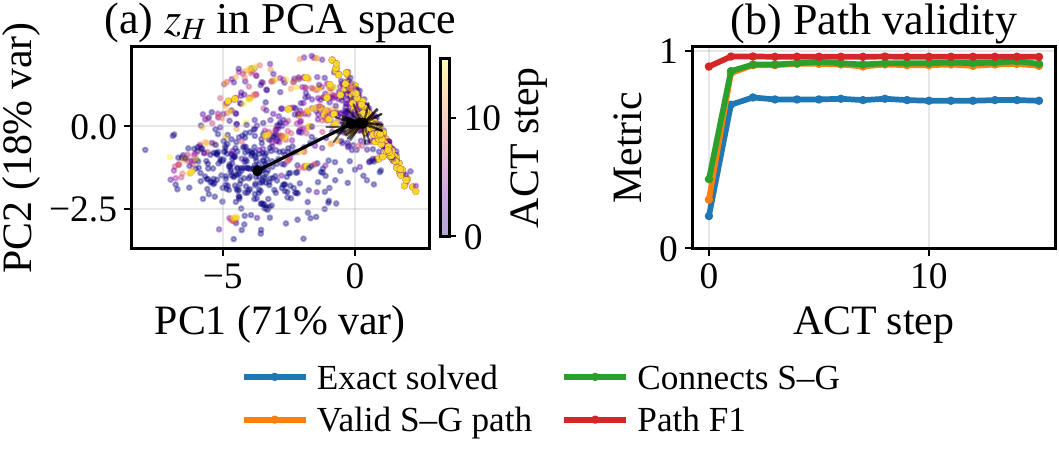}
\caption{Maze $\zH$ trajectory analysis over 300 puzzles. \textbf{(a)}~Trajectories projected onto the first two principal components. PC1 explains $70.6\%$ of the variance and PC2 explains $17.7\%$; the trajectories are predominantly linear and PC1-dominated, in contrast to the convergence-versus-wandering pattern observed on Sudoku. Dots denote step~0 and crosses denote step~15. \textbf{(b)}~Per-step path metrics: valid start--goal paths increase from $24.7\%$ at step~0 to $88.7\%$ after the first recurrent update and subsequently plateau near $93\%$, quantifying the early saturation in Maze.}
\label{fig:maze_trajectories}
\end{figure}

The Maze trajectories occupy a lower-dimensional manifold relative to Sudoku. Across $300$ puzzles, the first principal component explains $70.6\%$ of the variance, while both components together explain $88.3\%$ (bootstrap confidence interval $[86.3,89.8]$), with a mean consecutive state cosine similarity of $0.995$. Figure~\ref{fig:maze_trajectories} shows that the decoded predictions exhibit the same early saturation. Valid start-goal paths increase from $24.7\%$ at step~0 to $88.7\%$ after the first recurrent update and subsequently plateau near $93\%$. Maze therefore performs one dominant corrective update followed by comparatively small adjustments, in contrast to the gradual refinement across the ACT trajectory in Sudoku.



\subsection{Probe Targets}
\label{app:probes}

This appendix catalogs the full probe-target definitions for the probe study of Section~\ref{sec:finding2}; the corresponding sweep results appear in Appendix~\ref{app:full_probes}.  For every task, we cache HRM activations at recurrent steps and train independent probes of $\zH$ and $\zL$ for each target across segments. We train linear probe because their learned weights define directions that can be removed by projection; MLP probes are used as an expressivity check.

\paragraph{Sudoku.} For Sudoku, we consider the following targets for $\zH$ and $\zL$:
\begin{itemize}[leftmargin=*]
    \item \texttt{cell\_digit}: A ten-way multiclass target giving the digit currently encoded in a cell, $0$ for blank, $1-9$ otherwise. 
    \item \texttt{is\_given}: A binary flag of whether a cell was given in the input puzzle.
    \item \texttt{per\_cell\_correct}: A binary flag of whether the digit currently in the cell matches the label. 
    \item \texttt{violated\_in\_row}: A binary flag of whether the current digit in a cell appears more than once in its row.
    \item \texttt{violated\_in\_col}:A binary flag of whether the current digit in a cell appears more than once in its column.
    \item \texttt{violated\_in\_box}: A binary flag of whether the current digit in a cell appears more than once in its sub-grid.
\end{itemize}
All Sudoku probe rows use five seeds $\{0,1,2,3,4\}$, with $32{,}400$ train cells and $8{,}100$ validation cells.


\paragraph{Maze.}  Maze probes are split into global path probes and local cell probes. The global targets are 
\begin{itemize}[leftmargin=*]
    \item \texttt{exact\_solved}: A binary flag of whether the decoded grid matches the ground truth cell-for-cell. 
    \item \texttt{connects\_start\_goal}: A binary flag of whether source $S$ and goal $G$ cells connected by a path in the decoded grid.
    \item \texttt{valid\_sg\_path}: A binary flag of whether there is a path between start and goal and no path cell falls on a wall or has more than two path neighbors.
    \item \texttt{valid\_optimal\_path}: A binary flag of whether there is a valid path and the predicted route has exactly as many cells as the ground truth.
    \item \texttt{path\_f1}: F1 score between true and predicted path cells.
    \item \texttt{path\_jaccard}: Jaccard score between true and predicted path cells.
    \item \texttt{wall\_path\_rate}: Fraction of predicted path cells that land on wall cells.
    \item \texttt{path\_length\_ratio}: Fraction of predicted path cell count and optimal path cell count.
\end{itemize}
The local targets are 
\begin{itemize}[leftmargin=*]
    \item \texttt{on\_optimal\_path}: A binary flag indicating if a cell lies on ground truth shortest route.
    \item \texttt{is\_wall}: A binary flag indicating if the cell is a wall.
    \item \texttt{is\_free}: A binary flag indicating if the cell is free.
    \item \texttt{is\_dead\_end}: A binary flag indicating if a cell has exactly one passable neighbor.
    \item \texttt{is\_junction}: A binary flag indicating if a cell has two or more passable neighbors.
    \item \texttt{is\_corner\_on\_path}: A binary flag indicating if a cell has at least one vertical and horizontal neighbor.
    \item \texttt{off\_path\_passable}: A binary flag indicating if a cell is passable but not on the optimal route. 
    \item \texttt{near\_start\_5}: A binary flag indicating whether a cell is at most 5 passable cells away from the start cell.
    \item \texttt{near\_goal\_5}: A binary flag indicating whether a cell is at most 5 passable cells away from the goal cell.
    \item \texttt{num\_passable\_neighbors}: An integer indicating the number of passable neighbors.
    \item \texttt{distance\_to\_start\_norm}: The distance to the start cell normalized with the largest finite distance in the maze. 
    \item \texttt{distance\_to\_goal\_norm}: The distance to the goal cell normalized with the largest finite distance in the maze.
\end{itemize}
The Maze run uses puzzle-disjoint train/test splits over five seeds; global probes use $800$ training mazes and $200$ test mazes, while local probes use sampled positions with $102{,}400$ train cells and $25{,}600$ test cells.

\paragraph{ARC-AGI-2.}  ARC probes are representation-level checks rather than primary causal evidence. We evaluate the following per-cell ARC targets
\begin{itemize}[leftmargin=*]
    \item \texttt{per\_cell\_correct}: A binary flag of whether the token decoded at the cell matches the label. 
    \item \texttt{input\_inside\_grid}: A binary flag of whether the cell lies inside the rectangle of the input grid rather than in the surrounding padding.
    \item \texttt{output\_inside\_grid}: A binary flag of whether the cell lies inside the output grid rectangle.
    \item \texttt{color\_changed}: A binary flag of whether the cell holds a color in both the input and the output and the two differ, i.e.\ the
      transformation touched this cell.
    \item \texttt{same\_as\_input}: A binary flag of whether the cell holds the same color in the input and the output, i.e.\ the cell was copied.
    \item \texttt{is\_object\_boundary}: A binary flag of whether the cell is an input color cell $4$-adjacent to a cell of a different color.
    \item \texttt{input\_color}: A 10-class target giving the color of a cell in the input grid.
    \item \texttt{output\_color}: A 10-class target indicating the color of a cell in the target output grid.
\end{itemize}
as well as the following per-grid targets
\begin{itemize}[leftmargin=*]
    \item \texttt{exact\_solved}: A binary flag of whether the decoded grid matches the label at every scored cell. 
    \item \texttt{shape\_correct}: A binary flag of whether the decoded color region has the same height and width as the label. 
    \item \texttt{num\_input\_colors}: The number of distinct colors present in the input grid.  
    \item \texttt{num\_output\_colors}: The number of distinct colors present in the output grid.
\end{itemize}

\FloatBarrier

\subsection{Full Probe Results and Sweeps}
\label{app:full_probes}

\begin{figure*}[!t]
\centering
\includegraphics[width=0.95\textwidth]{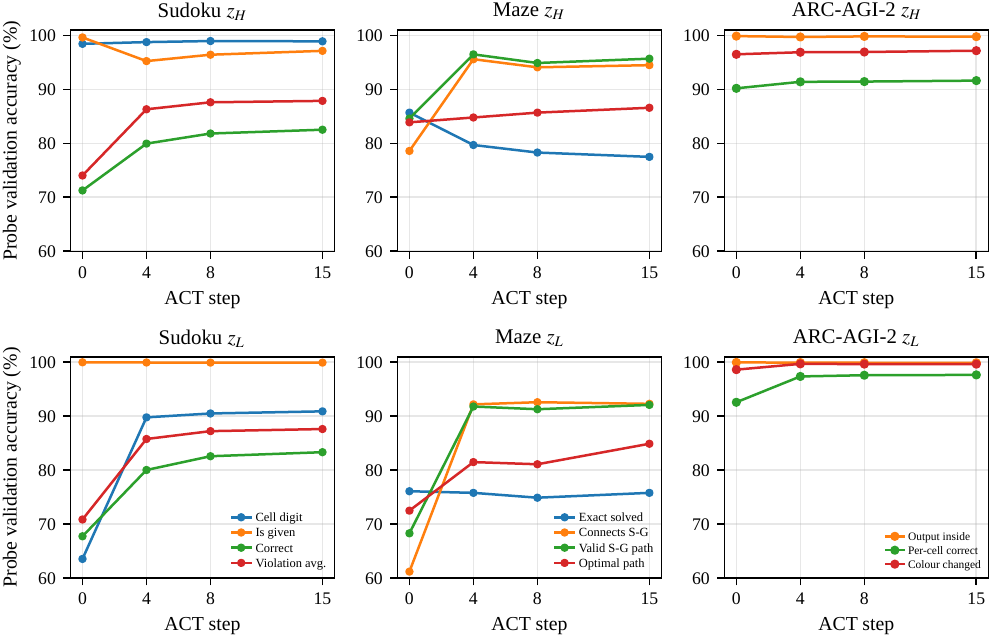}
\caption{Representative linear-probe sweeps for the three tasks. Rows correspond to $\zH$ and $\zL$. Sudoku probes report given-cell identity, per-cell correctness, and the mean of row-, column-, and box-violation probes. Maze probes report validation accuracy for four global path targets. ARC-AGI-2 shows output-grid membership, per-cell correctness, and color change. Task variables are readable from both $\zH$ and $\zL$, even when the causal interventions in Finding~\ref{sec:finding1} assign different roles to the two states.}
\label{fig:probe_sweeps_sudoku_maze}
\end{figure*}

This section reports the complete probe sweeps for Section~\ref{sec:finding2}. Figure~\ref{fig:probe_sweeps_sudoku_maze} compares all three tasks across fixed steps. The complete target results can be found in Table~\ref{tab:sudoku_probe_full}, ~\ref{tab:maze_probe_full}, and ~\ref{tab:arc_probe_full}.  

\paragraph{Sudoku.}
Table~\ref{tab:sudoku_probe_full} reports the per-step sweeps for probe accuracy. Sudoku separates static from dynamic variables. Static features are decodable from $\zH$ at step~0, with \texttt{cell\_digit} at $98.5\%$ and \texttt{is\_given} at $99.6\%$, and they remain flat thereafter. Dynamic variables instead improve over recurrence: \texttt{per\_cell\_correct} rises from $71.3\%$ and the mean violation probe from $74.1\%$, each gaining $11$-$14$ points by step~15. 

\paragraph{Maze.}
Table~\ref{tab:maze_probe_full} reports both global and local feature results. Across steps, start–goal connectivity rises from $78.6\%$ to $94.5\%$ in
$\zH$ and from $61.2\%$ to $92.3\%$ in $\zL$. This improvement is target specific since \texttt{exact\_solved} does not improve. Local layout
variables are accessible from the earliest state: wall and free-cell status
are decoded perfectly, and optimal-path membership exceeds $95\%$ for both
streams.

\paragraph{ARC-AGI-2}
ARC-AGI-2 extends the readout result to heterogeneous grid transformations. Output and input grid membership are almost fully decodable, while \texttt{per\_cell\_correct} rises from $90.2\%$ to $91.6\%$ in $\zH$ and
from $92.6\%$ to $97.7\%$ in $\zL$. Table~\ref{tab:arc_probe_full} additionally reports other color, boundary, and per-cell targets.  

\begin{table*}[!t]
\centering
\caption{Complete Sudoku linear-probe endpoint results. Values are validation accuracy (\%), reported as mean $\pm$ standard deviation over five seeds. Each probe uses $32{,}400$ training cells and $8{,}100$ validation cells.}
\label{tab:sudoku_probe_full}
\vskip 0.1in
\small
\begin{tabular}{@{}lcccc@{}}
\toprule
\textbf{Target} & $\zH^{(0)}$ & $\zH^{(15)}$ & $\zL^{(0)}$ & $\zL^{(15)}$ \\
\midrule
\texttt{cell\_digit} & $98.5 \pm 0.1$ & $98.9 \pm 0.1$ & $63.6 \pm 0.4$ & $90.9 \pm 0.5$ \\
\texttt{is\_given} & $99.6 \pm 0.1$ & $97.2 \pm 0.5$ & $100.0 \pm 0.0$ & $99.9 \pm 0.0$ \\
\texttt{per\_cell\_correct} & $71.3 \pm 0.3$ & $82.5 \pm 1.2$ & $67.7 \pm 0.6$ & $83.3 \pm 1.5$ \\
\texttt{violated\_in\_row} & $74.1 \pm 0.6$ & $88.3 \pm 0.6$ & $71.4 \pm 0.8$ & $88.2 \pm 0.7$ \\
\texttt{violated\_in\_col} & $74.1 \pm 0.6$ & $87.8 \pm 0.6$ & $70.9 \pm 0.6$ & $87.6 \pm 0.8$ \\
\texttt{violated\_in\_box} & $74.0 \pm 0.5$ & $87.5 \pm 0.7$ & $70.3 \pm 0.2$ & $87.1 \pm 0.8$ \\
\bottomrule
\end{tabular}
\end{table*}

\begin{table*}[!t]
\centering
\caption{Complete Maze linear-probe endpoint results. Categorical targets report validation accuracy (\%) as mean $\pm$ standard deviation over five puzzle-disjoint seeds. Global probes use $800/200$ train/test mazes; local probes use $102{,}400/25{,}600$ train/test cells. \textnormal{n/r} denotes continuous targets whose $R^2$ is explicitly marked unreliable because the evaluation target is nearly constant.}
\label{tab:maze_probe_full}
\vskip 0.1in
\scriptsize
\begin{tabular}{@{}lcccc@{}}
\toprule
\textbf{Target} & $\zH^{(0)}$ & $\zH^{(15)}$ & $\zL^{(0)}$ & $\zL^{(15)}$ \\
\midrule
\multicolumn{5}{@{}l}{\textit{Global path targets}} \\
\texttt{exact\_solved} & $85.7 \pm 3.3$ & $77.5 \pm 1.5$ & $76.1 \pm 3.1$ & $75.8 \pm 3.7$ \\
\texttt{connects\_start\_goal} & $78.6 \pm 3.9$ & $94.5 \pm 2.6$ & $61.2 \pm 3.2$ & $92.3 \pm 3.0$ \\
\texttt{valid\_sg\_path} & $84.6 \pm 2.2$ & $95.7 \pm 2.1$ & $68.3 \pm 4.6$ & $92.1 \pm 1.7$ \\
\texttt{valid\_optimal\_path} & $83.9 \pm 3.4$ & $86.6 \pm 1.8$ & $72.5 \pm 2.6$ & $84.9 \pm 1.7$ \\
\texttt{path\_f1} & n/r & n/r & n/r & n/r \\
\texttt{path\_jaccard} & n/r & n/r & n/r & n/r \\
\texttt{wall\_path\_rate} & n/r & n/r & n/r & n/r \\
\texttt{path\_length\_ratio} & n/r & n/r & n/r & n/r \\
\midrule
\multicolumn{5}{@{}l}{\textit{Local cell targets}} \\
\texttt{on\_optimal\_path} & $98.2 \pm 0.2$ & $98.8 \pm 0.1$ & $95.7 \pm 0.2$ & $98.9 \pm 0.1$ \\
\texttt{is\_wall} & $100.0 \pm 0.0$ & $100.0 \pm 0.0$ & $100.0 \pm 0.0$ & $100.0 \pm 0.0$ \\
\texttt{is\_free} & $100.0 \pm 0.0$ & $100.0 \pm 0.0$ & $100.0 \pm 0.0$ & $100.0 \pm 0.0$ \\
\texttt{is\_dead\_end} & $96.6 \pm 0.2$ & $96.6 \pm 0.1$ & $97.7 \pm 0.3$ & $97.1 \pm 0.3$ \\
\texttt{is\_junction} & $86.9 \pm 0.5$ & $84.4 \pm 1.0$ & $93.2 \pm 0.1$ & $91.0 \pm 0.2$ \\
\texttt{is\_corner\_on\_path} & $93.2 \pm 0.2$ & $95.5 \pm 0.2$ & $92.8 \pm 0.1$ & $97.5 \pm 0.1$ \\
\texttt{off\_path\_passable} & $98.3 \pm 0.1$ & $98.7 \pm 0.2$ & $95.9 \pm 0.1$ & $98.9 \pm 0.1$ \\
\texttt{near\_start\_5} & $97.9 \pm 0.1$ & $97.9 \pm 0.1$ & $98.2 \pm 0.1$ & $98.2 \pm 0.1$ \\
\texttt{near\_goal\_5} & $98.0 \pm 0.1$ & $98.0 \pm 0.1$ & $98.0 \pm 0.1$ & $98.2 \pm 0.1$ \\
\texttt{num\_passable\_neighbors} & n/r & n/r & n/r & n/r \\
\texttt{distance\_to\_start\_norm} & n/r & n/r & n/r & n/r \\
\texttt{distance\_to\_goal\_norm} & n/r & n/r & n/r & n/r \\
\bottomrule
\end{tabular}
\end{table*}

\begin{table*}[!t]
\centering
\caption{Complete ARC-AGI-2 linear-probe endpoint results. Reportable categorical targets give validation accuracy (\%) as mean $\pm$ standard deviation over five seeds, except \texttt{shape\_correct} at step~15 ($n=4$ valid seeds). Dashes denote one-class step-0 targets for which a classifier cannot be fit. \textnormal{n/r} denotes near-constant continuous targets with unreliable $R^2$.}
\label{tab:arc_probe_full}
\vskip 0.1in
\scriptsize
\begin{tabular}{@{}lcccc@{}}
\toprule
\textbf{Target} & $\zH^{(0)}$ & $\zH^{(15)}$ & $\zL^{(0)}$ & $\zL^{(15)}$ \\
\midrule
\multicolumn{5}{@{}l}{\textit{Per-cell targets}} \\
\texttt{per\_cell\_correct} & $90.2 \pm 0.6$ & $91.6 \pm 0.5$ & $92.6 \pm 0.4$ & $97.7 \pm 0.3$ \\
\texttt{input\_inside\_grid} & $99.9 \pm 0.1$ & $99.7 \pm 0.2$ & $100.0 \pm 0.0$ & $99.9 \pm 0.0$ \\
\texttt{output\_inside\_grid} & $99.9 \pm 0.1$ & $99.8 \pm 0.1$ & $100.0 \pm 0.0$ & $99.9 \pm 0.0$ \\
\texttt{colour\_changed} & $96.5 \pm 0.2$ & $97.2 \pm 0.3$ & $98.6 \pm 0.1$ & $99.7 \pm 0.1$ \\
\texttt{same\_as\_input} & $96.0 \pm 0.2$ & $97.6 \pm 0.3$ & $98.7 \pm 0.0$ & $99.7 \pm 0.1$ \\
\texttt{is\_object\_boundary} & $85.0 \pm 0.5$ & $82.0 \pm 0.4$ & $98.2 \pm 0.1$ & $97.6 \pm 0.1$ \\
\texttt{input\_colour} & $91.5 \pm 0.3$ & $81.4 \pm 0.8$ & $99.9 \pm 0.1$ & $99.9 \pm 0.1$ \\
\texttt{output\_colour} & $81.2 \pm 0.8$ & $88.6 \pm 1.5$ & $78.5 \pm 0.5$ & $86.5 \pm 0.7$ \\
\midrule
\multicolumn{5}{@{}l}{\textit{Per-grid targets}} \\
\texttt{exact\_solved} & --- & $92.0 \pm 3.5$ & --- & $87.8 \pm 3.3$ \\
\texttt{shape\_correct} & --- & $97.7 \pm 3.9$ & --- & $95.7 \pm 2.5$ \\
\texttt{num\_input\_colours} & n/r & n/r & n/r & n/r \\
\texttt{num\_output\_colours} & n/r & n/r & n/r & n/r \\
\bottomrule
\end{tabular}
\end{table*}

\subsubsection{Linear Probe vs. Nonlinear Probe}
\label{app:mlp_probes_main}
\begin{figure}[t]
\centering
\includegraphics[width=.9\columnwidth]{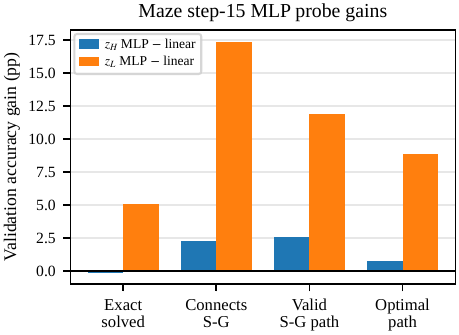}
\caption{MLP-minus-linear probe gains on Maze at step~15. Nonlinear probes provide modest gains for $\zH$ and larger gains for several $\zL$ path targets, particularly start--goal connectivity and valid-path status. These gains indicate nonlinear accessibility rather than causal relevance.}
\label{fig:maze_mlp_gain_step15}
\end{figure}

\begin{figure}[!htbp]
\centering
\includegraphics[width=.9\columnwidth]{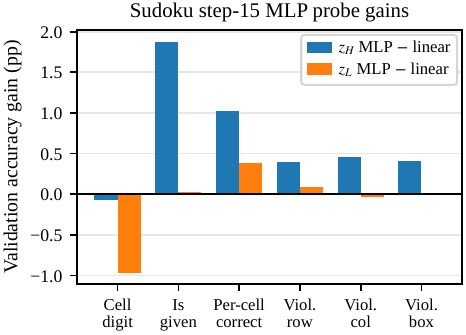}
\caption{MLP-minus-linear probe gains on Sudoku at step~15. No target
gains more than $1.9$ points from a nonlinear readout, and the mean gain is
$+0.68$ for $\zH$ and $-0.08$ for $\zL$. The Sudoku constraint variables are
therefore predominantly linearly accessible from both streams.}
\label{fig:sudoku_mlp_gain_step15}
\end{figure}

We train two-layer MLP probes with 512 dimensions and ReLU activation in the first layer, and 256 dimensions and ReLU activation in the second layer, using the same train/test splits as the linear probes. On Sudoku (Figure~\ref{fig:sudoku_mlp_gain_step15}), the mean gain across the six step-15 targets is below $1\%$, so the reported Sudoku constraint variables are therefore predominantly linearly accessible. On Maze, the gains separate by $\zH$ and $\zL$ (Figure~\ref{fig:maze_mlp_gain_step15}); $\zH$ gains are comparable to linear counterparts, while $\zL$ gains are up to $17.4$ points, concentrated on start-goal connectivity. Some Maze path information is therefore more accessible through a nonlinear readout from $\zL$. This result does not alter the causal conclusion in Section~\ref{sec:finding2}, since an MLP measures nonlinear accessibility but does not define a single linear direction whose removal can be interpreted as a targeted intervention.




\subsection{Interpretable Geometry of Probe Weight Vectors}
\label{app:geometry}

Here, we examine whether the learned probe weights define structured readout directions rather than unrelated classifiers. For each target, we unit-normalize the learned weight vector, compute pairwise cosine similarities, and apply PCA to the resulting set of directions.

\paragraph{Sudoku.}
In Sudoku (Table~\ref{tab:geometry_sudoku}), the row, column, and sub-grid violation directions cluster together (mean cosine $0.92$--$0.97$), while the correctness direction is oriented oppositely ($-0.33$ to $-0.50$; Table~\ref{tab:geometry_sudoku}). Accordingly, PC1 explains $93.3$--$97.2\%$ of the variance. The exact probe directions vary across seeds (same-target agreement $\approx 0.71$). However, the three violation probes remain closely aligned with one another (cosine $\approx 0.95$), suggesting that they decode a shared constraint violation direction. In contrast, the \texttt{is\_given} is nearly orthogonal to these directions (cosines $0.16$--$0.18$). 

\begin{table}[!htbp]
\centering
\caption{Sudoku probe geometry across recurrent steps. V-V is the mean cosine
between violation directions; C-V is the mean cosine between correctness and
violation directions; PC1 is computed from the four centered, unit-normalized
directions; Rel. is mean same-target agreement across seeds. Values are means
over five seeds. Random-direction nulls: $|\cos|=0.035$ and PC1 $=35.9\%$.}
\label{tab:geometry_sudoku}
\vskip 0.1in
\small
{\setlength{\tabcolsep}{4pt}
\begin{tabular}{@{}llrrrrr@{}}
\toprule
& & \multicolumn{5}{c}{\textbf{Step}} \\
\cmidrule(l){3-7}
& & $\mathbf{0}$ & $\mathbf{4}$ & $\mathbf{8}$ & $\mathbf{12}$ & $\mathbf{15}$ \\
\midrule
$\zH$ & V--V     & $+0.921$ & $+0.947$ & $+0.953$ & $+0.953$ & $+0.953$ \\
$\zH$ & C--V     & $-0.485$ & $-0.369$ & $-0.346$ & $-0.342$ & $-0.330$ \\
$\zH$ & PC1 (\%) & $93.3$   & $95.0$   & $95.5$   & $95.5$   & $95.4$ \\
$\zH$ & Rel.     & $0.74$   & $0.73$   & $0.72$   & $0.71$   & $0.71$ \\
\midrule
$\zL$ & V--V     & $+0.966$ & $+0.964$ & $+0.970$ & $+0.967$ & $+0.969$ \\
$\zL$ & C--V     & $-0.499$ & $-0.366$ & $-0.385$ & $-0.383$ & $-0.389$ \\
$\zL$ & PC1 (\%) & $97.0$   & $96.6$   & $97.2$   & $96.9$   & $97.1$ \\
$\zL$ & Rel.     & $0.73$   & $0.70$   & $0.71$   & $0.71$   & $0.71$ \\
\bottomrule
\end{tabular}
}
\end{table}

\paragraph{Maze.}
Maze shows that probe accuracy does not guarantee a reproducible direction (Table~\ref{tab:geometry_maze}). At $\zH$ step~15, only the wall, free cell, optimal path, and off-path passable directions are stable across seeds (cosine: $0.47$-$0.79$). Several other probes reach $77.8\%$-$84.9\%$ accuracy but recover unstable directions. PC1 explains $55.8\%$ of the variance. Maze probe geometry is therefore structured but multidimensional, with directions generally more stable in $\zL$. 

\begin{table*}[!t]
\centering
\caption{Maze probe geometry at step~15. Panel A reports validation accuracy
from the geometry refits and same-target seed reliability. Panel B reports
pairwise cosines among the four stable directions and PC1 for the stable set
and all nine directions. Values are means over five seeds.}
\label{tab:geometry_maze}
\vskip 0.1in
\small

\textbf{Panel A: per-target accuracy and direction reliability}
\vskip 0.05in
{\setlength{\tabcolsep}{6pt}
\begin{tabular}{@{}lrrrr@{}}
\toprule
& \multicolumn{2}{c}{$\zH$} & \multicolumn{2}{c}{$\zL$} \\
\cmidrule(lr){2-3} \cmidrule(l){4-5}
\textbf{Target} & \textbf{Acc.\ (\%)} & \textbf{Rel.} &
\textbf{Acc.\ (\%)} & \textbf{Rel.} \\
\midrule
\texttt{is\_wall}            & $100.0$ & $0.79$ & $100.0$ & $0.84$ \\
\texttt{is\_free}            & $100.0$ & $0.76$ & $100.0$ & $0.86$ \\
\texttt{on\_optimal\_path}   & $98.6$  & $0.77$ & $98.9$  & $0.84$ \\
\texttt{off\_path\_passable} & $98.5$  & $0.47$ & $98.4$  & $0.85$ \\
\midrule
\texttt{is\_corner\_on\_path} & $81.0$ & $0.19$ & $88.2$ & $0.81$ \\
\texttt{is\_junction}         & $77.8$ & $0.07$ & $86.1$  & $0.70$ \\
\texttt{near\_goal\_5}        & $84.9$ & $0.06$ & $89.8$  & $0.34$ \\
\texttt{near\_start\_5}       & $83.0$ & $0.06$ & $89.2$  & $0.29$ \\
\texttt{is\_dead\_end}        & $83.5$ & $0.02$ & $89.5$  & $0.22$ \\
\bottomrule
\end{tabular}
}

\vskip 0.12in
\textbf{Panel B: geometry of the four stable directions}
\vskip 0.05in
{\setlength{\tabcolsep}{5pt}
\begin{tabular}{@{}lrrrrrrrr@{}}
\toprule
& \multicolumn{6}{c}{\textbf{Pairwise cosine}} &
  \multicolumn{2}{c}{\textbf{PC1 (\%)}} \\
\cmidrule(lr){2-7} \cmidrule(l){8-9}
\textbf{State} &
\textbf{wall--free} & \textbf{wall--path} & \textbf{wall--off} &
\textbf{free--path} & \textbf{free--off} & \textbf{path--off} &
\textbf{4 dir.} & \textbf{9 dir.} \\
\midrule
$\zH$ & $-0.694$ & $-0.432$ & $-0.184$ & $+0.389$ & $+0.218$ & $-0.113$ & $55.8$ & $24.5$ \\
$\zL$ & $-0.723$ & $-0.463$ & $-0.190$ & $+0.380$ & $+0.243$ & $-0.165$ & $56.0$ & $25.0$ \\
\midrule
Null  & $0.035$ & $0.035$ & $0.035$ & $0.035$ & $0.035$ & $0.035$ & $35.9$ & $14.9$ \\
\bottomrule
\end{tabular}
}
\end{table*}

\paragraph{ARC-AGI-2}
ARC probe directions share meaningful structure but do not collapse onto a small number of common axes (Table~\ref{tab:geometry_arc}). In particular, \texttt{color\_changed} and \texttt{same\_as\_input} point in strongly opposing directions across streams and steps ($-0.780$ to $-0.916$). PC1 explains $36.9\%$ - $52.0\%$ of the variance. However, four components are required to explain $90\%$ of the variation. This indicates that ARC probe directions remain multidimensional. 

\begin{table}[!t]
\centering
\caption{ARC-AGI-2 geometry for six probe directions. Changed,
copied, correct, boundary, in-grid, and eos denote \texttt{colour\_changed},
\texttt{same\_as\_input}, \texttt{per\_cell\_correct},
\texttt{is\_object\_boundary}, \texttt{input\_inside\_grid}, and
\texttt{is\_eos}.}
\label{tab:geometry_arc}
\vskip 0.1in
\small
{\setlength{\tabcolsep}{4pt}
\begin{tabular}{@{}lrrrr@{}}
\toprule
& \multicolumn{2}{c}{$\zH$} & \multicolumn{2}{c}{$\zL$} \\
\cmidrule(lr){2-3} \cmidrule(l){4-5}
\textbf{Direction pair} & \textbf{step 0} & \textbf{step 15} &
\textbf{step 0} & \textbf{step 15} \\
\midrule
changed--copied    & $-0.873$ & $-0.780$ & $-0.916$ & $-0.883$ \\
correct--copied    & $+0.573$ & $+0.328$ & $+0.274$ & $+0.020$ \\
correct--changed   & $-0.551$ & $-0.139$ & $-0.234$ & $+0.016$ \\
changed--boundary  & $+0.386$ & $+0.256$ & $+0.079$ & $+0.057$ \\
correct--in-grid   & $+0.028$ & $+0.257$ & $+0.143$ & $+0.257$ \\
in-grid--eos       & $-0.232$ & $-0.368$ & $-0.292$ & $-0.297$ \\
\midrule
PC1 (\%)           & $52.0$ & $40.2$ & $38.7$ & $36.9$ \\
PC1--2 (\%)        & $75.5$ & $67.8$ & $64.9$ & $63.4$ \\
\bottomrule
\end{tabular}
}
\end{table}

Overall, the probes recover semantically organized readout geometry, but this does not establish that the model computes along these directions. PCA here describes only a small set of probe weights, not the dimensionality of activation space.

\subsection{SAE Training and Sweep Results}
\label{app:sae_sweeps}

For Sudoku and Maze, we perform a grid search over dictionary sizes $D_{\mathrm{SAE}} \in \{1024,2048,4096,8192\}$ and sparsity penalties $\lambda_{\mathrm{SAE}} \in \{0.003,0.01,0.03\}$. We find that the linear probes achieve similar performance across different configurations. Thus, we choose $D_{\mathrm{SAE}} = 2048$, $\lambda_{\mathrm{SAE}} = 0.01$ for all experiments. We choose the top-50 features from SAE. 

\begin{table}[tbp]
\centering
\caption{SAE causal effects used in Figure~\ref{fig:readout_vs_causality}(b). Mean metric change on the primary metric of each task (i.e., cell accuracy for Sudoku, valid path rate for Maze, color-cell accuracy for ARC-AGI-2). $p$ tests top-50 $=$ random-50.}
\label{tab:sae_rep_app}
\vskip 0.1in
\small
\begin{tabular}{@{}lrrrr@{}}
\toprule
\textbf{Task} & $n$ & \textbf{Top-50} & \textbf{Rand.-50} & $p$ \\
\midrule
Sudoku    & 300 & $-3.63$ & $-3.42$ & 0.22 \\
Maze      & 300 & $-2.99$ & $-3.35$ & 0.18 \\
ARC-AGI-2 & 150 & $+1.27$ & $+1.28$ & 0.94 \\
\bottomrule
\end{tabular}
\end{table}

\paragraph{Top-$k$ sensitivity.}
On Sudoku, varying $k$ does not produce a stable top-feature advantage (Table~\ref{tab:sae_topk_app}). The top set is modestly more damaging at $k = 16$, but the gap vanishes or reverses for larger $k$.

\begin{table}[tbp]
\centering
\caption{Sudoku top-$k$ SAE sensitivity. Values are mean $\Delta$cell accuracy (\%).}
\label{tab:sae_topk_app}
\vskip 0.1in
\small
\begin{tabular}{@{}rrrrr@{}}
\toprule
$k$ & \textbf{Top-$k$} & \textbf{Random-$k$} & \textbf{Top--Random} & \textbf{$p$} \\
\midrule
16 & $-3.08$ & $-2.13$ & $-0.95$ & $2.9\times10^{-4}$ \\
32 & $-12.12$ & $-12.37$ & $+0.25$ & 0.45 \\
50 & $-3.63$ & $-3.42$ & $-0.21$ & 0.23 \\
64 & $-3.92$ & $-4.44$ & $+0.52$ & 0.085 \\
\bottomrule
\end{tabular}
\end{table}

\paragraph{Dead feature analysis.}
We perform a post-hoc analysis on dead features, i.e., features that are not activated during inference. We calculate the activation rate of activated features with respect to all features. In our analysis, we found 704 dead features, $1{,}095$ features with $<0.1\%$ activation rate, 22 features with 0.1--1\% activation rate, 65 features with 1--10\% activation rate, 124 features with 10--50\% activation rate, and 38 with $>50\%$ activation rate.  The dictionary is therefore sparse but not semantically localized.

\subsection{SAE Top Features}
\label{app:sae_features}
To test whether individual SAE features correspond to individual constraints, we encode $\zH$ activations with an SAE of dictionary size $D = 2048$ and sparsity penalty $\lambda = 0.01$. We calculate the correlation between each feature activation and each per-cell binary targets:

\begin{itemize}[leftmargin=*,nosep]
    \item \texttt{per\_cell\_correct}: A binary flag of whether the digit currently in the cell matches the label. 
    \item \texttt{is\_given}: A binary flag of whether a cell was given in the input puzzle.
    \item \texttt{violated\_in\_row/col/box}: Binary flags that check whether the current digit in a cell appears more than once in its row, column or box (sub--grid), respectively. 
    \item \texttt{is\_naked\_single}: A binary flag that checks whether there is only one possible digit that can be entered in a cell.
    \item \texttt{is\_hidden\_single\_row/col/box}: Binary flags that check whether a digit can be placed in this cell but cannot be placed in any other row/column/box of that cell. 
\end{itemize} 
For each binary target, we treat the learned weight vector as a direction in activation space and analyze the resulting set of directions using cosine similarity and PCA. In Sudoku, the row-, column-, and box-violation directions cluster together with correctness oriented inversely, and Maze shows an analogous grouping among walls, free cells, paths, and proximity to the start or goal.
For every feature, we record its largest correlation with a target $|r_{\text{best}}|$ and the largest correlation over the remaining nine targets $r_{\text{next}}$. We call a feature mono-semantic if $|r_{\text{best}}| > 2|r_{\text{next}}|$. Table~\ref{tab:sae_top_features} lists the most target-correlated Sudoku SAE features. No feature satisfies $|r_{\text{best}}| > 2|r_{\text{next}}|$, with most leading features having a ratio between $1.0$ and $1.3$. The dictionary, therefore, does not decompose $\zH$ into constraint-specific features. 

\begin{table}[!t]
\centering
\caption{Most-specialized Sudoku SAE features ($D = 2048$, $\lambda = 0.01$). $r_{\text{best}}$ is the Pearson correlation between the feature and the best target; $r_{\text{next}}$ is the largest absolute correlation between the feature and any other target.}
\label{tab:sae_top_features}
\vskip 0.1in
\small
\begin{tabular}{@{}lllrr@{}}
\toprule
\textbf{Feat.} & \textbf{Best target} & \textbf{Peak step} & $r_{\text{best}}$ & $r_{\text{next}}$ \\
\midrule
263  & box violation       & 0  & $+0.41$ & 0.40 \\
597  & is-given            & 13 & $+0.39$ & 0.33 \\
1809 & is-given            & -- & $+0.39$ & 0.21 \\
1542 & per-cell correct    & 0  & $-0.37$ & 0.34 \\
654  & is-given            & -- & $+0.35$ & 0.31 \\
211  & is-given            & 2  & $-0.35$ & 0.33 \\
1317 & per-cell correct    & -- & $+0.33$ & 0.26 \\
2015 & per-cell correct    & 14 & $+0.33$ & 0.32 \\
2031 & box violation       & 0  & $+0.32$ & 0.32 \\
1715 & per-cell correct    & 1  & $-0.32$ & 0.28 \\
1576 & is-given            & -- & $-0.32$ & 0.23 \\
639  & is-given            & -- & $-0.31$ & 0.26 \\
1228 & per-cell correct    & 1  & $-0.31$ & 0.31 \\
302  & is-given            & 0  & $-0.31$ & 0.28 \\
\bottomrule
\end{tabular}
\end{table}

\subsection{Details of Training HRM with BPTT}
\label{app:bptt}

We train two HRM models from scratch for Sudoku using the same data, optimizer schedule, and random seed: a vanilla HRM using the one-step gradient approximation, and a BPTT variant with full within-step backpropagation, while still detaching the latent states from the computation graph across segments.  For each model, we train a SAE of the same architecture and select the Pareto-best configuration, i.e., dictionary size $D_{\mathrm{SAE}} = 2048$ and sparsity penalty $\lambda_{\mathrm{SAE}} = 0.003$, for causal ablation. We use random seed 42 for the evaluation. Both models are evaluated on the same $200$ test puzzles, and top-50 SAE features are extracted. Probe-direction conditions are not reported in this study because the corresponding aggregate rows have zero samples.

\begin{table}[H]
\centering
\caption{BPTT SAE causal-ablation details. Values are mean $\Delta$cell accuracy (\%).}
\label{tab:bptt_app}
\vskip 0.1in
\small
\begin{tabular}{@{}lcccc@{}}
\toprule
\textbf{Model} & \textbf{Top-50} & \textbf{Random-50} & \textbf{paired $p$} & \textbf{Cohen's $d$} \\
\midrule
Stock HRM & $-1.59$ & $-1.37$ & 0.39 & $-0.06$ \\
BPTT HRM  & $-29.64$ & $-29.63$ & 0.93 & $-0.01$ \\
\bottomrule
\end{tabular}

\par\medskip\normalsize
\caption{Best feature--target correlations in the BPTT training-regime check. ``Spec.'' counts features satisfying the specialization criterion $|r_{\text{best}}|>2|r_{\text{next}}|$; neither model develops highly specialized symbolic features.}
\label{tab:bptt_specialization_app}
\vskip 0.1in
\small
\begin{tabular}{@{}llllc@{}}
\toprule
\textbf{Model} & \textbf{Feat.} & \textbf{Target} & $r_{\text{best}}$ / $r_{\text{next}}$ & \textbf{Spec.} \\
\midrule
Stock HRM & 1619 & is-given & $-0.52$ / $0.36$ & 0 \\
BPTT HRM  & 92   & is-given & $-0.55$ / $0.47$ & 0 \\
\bottomrule
\end{tabular}
\end{table}

\FloatBarrier

BPTT increases the causal mass captured by the SAE basis, but not its localization: top-50 and random-50 ablations remain indistinguishable, and the most correlated features are still broadly mixed across symbolic targets. This supports the conclusion that the distributed-computation result is not merely an artifact of one-step gradient approximation.

\end{document}